\documentclass[10pt,twocolumn,letterpaper]{article}

\usepackage[pagenumbers]{cvpr} 

\usepackage{graphicx}
\usepackage{amsmath}
\usepackage{booktabs}
\usepackage{amssymb}
\usepackage{pifont}
\usepackage{colortbl}
\usepackage{marvosym}
\usepackage{booktabs, multirow} 
\usepackage{algorithm}
\usepackage{algpseudocode}
\usepackage{stfloats}
\usepackage{tikz}
\usepackage{xcolor}
\usepackage{pgfplots}
\usepackage{graphicx}

\usepackage{tcolorbox}
\definecolor{darkgreen}{RGB}{0, 128, 0}
\definecolor{darkred}{RGB}{196, 0, 0}
\definecolor{darkblue}{RGB}{64, 128, 196}

\usepackage{etoolbox, subcaption}
\usepackage[font=footnotesize,labelfont=bf,aboveskip=2pt,belowskip=-8pt]{caption}

\newdimen\abovecrulesep
\newdimen\belowcrulesep
\makeatletter
\patchcmd{\@@@cmidrule}{\aboverulesep}{\abovecrulesep}{}{}
\patchcmd{\@xcmidrule}{\belowrulesep}{\belowcrulesep}{}{}

\definecolor{demphcolor}{RGB}{144, 144, 144}
\definecolor{mygray}{gray}{0.4}
\definecolor{lightgray}{rgb}{0.9, 0.9, 0.9}

\newlength\savewidth

\newcommand{\tablestyle}[2]{\setlength{\tabcolsep}{#1}\renewcommand{\arraystretch}{#2}\centering\footnotesize}
\makeatletter\renewcommand\paragraph{\@startsection{paragraph}{4}{\z@}{.5em\@plus1ex\@minus.2ex}{-.5em}{\normalfont\normalsize\bfseries}}
\makeatother

\newcolumntype{C}[1]{>{\centering\arraybackslash}p{#1}}
\newcolumntype{R}[1]{>{\raggedleft\arraybackslash}p{#1}}
\newcolumntype{L}[1]{>{\raggedright\arraybackslash}p{#1}}

\usepackage{etoolbox}
\preto\align{\small}
\expandafter\preto\csname align*\endcsname{\small}
\preto\equation{\par\nobreak\small\noindent}

\usepackage{xcolor}
\definecolor{citecolor}{HTML}{0071bc}
\usepackage[pagebackref=false,breaklinks=true,letterpaper=true,colorlinks,citecolor=citecolor,bookmarks=false]{hyperref}
\usepackage{float}
\newcommand{\cmark}{\text{\ding{51}}} 
\newcommand{\xmark}{\text{\ding{55}}} 

\newcommand{\SHOW}{\textcolor{scolor}{S}\textcolor{hcolor}{h}\textcolor{ocolor}{o}\textcolor{wcolor}{w}UI}
\newcommand{\our}{ShowUI\xspace}
\newcommand{\baseshort}{Qwen2-VL-2B}
\newcommand{\base}{Qwen2-VL-2B}

\usepackage{graphicx}
\usepackage{subcaption}

\newcommand{\screenspot}{Screenspot}

\newcommand{\mindweb}{Mind2Web}

\definecolor{scolor}{RGB}{111,168,220}
\definecolor{hcolor}{RGB}{111,176,81}
\definecolor{ocolor}{RGB}{224,103,102}
\definecolor{wcolor}{RGB}{246,178,107}

\usepackage[capitalize]{cleveref}
\crefname{section}{Sec.}{Secs.}
\Crefname{section}{Section}{Sections}
\Crefname{table}{Table}{Tables}
\crefname{table}{Tab.}{Tabs.}

\def\cvprPaperID{475} 
\def\confName{CVPR}
\def\confYear{2024}

\definecolor{fuchsia}{rgb}{0.6, 0.33, 0.73}
\usepackage[pagebackref=false,breaklinks=true,letterpaper=true,colorlinks,citecolor=citecolor,bookmarks=false]{hyperref}

\definecolor{NiceBlue}{rgb}{0.11764705882352941, 0.5647058823529412, 1.0}
\definecolor{Gray}{gray}{0.9}
\hypersetup{
    colorlinks=citecolor,
    linkcolor=citecolor, 
}
\begin{document}

\title{\SHOW: One Vision-Language-Action Model for GUI Visual Agent}

\author{%
	Kevin Qinghong Lin$^1$,~
	Linjie Li$^2$,~
	Difei Gao$^1$,~
        Zhengyuan Yang$^2$,~ 
        \\
        {
        Shiwei Wu$^1$,~ 
        Zechen Bai$^1$,~ 
        Stan Weixian Lei$^1$,~ 
        Lijuan Wang$^2$,~
        Mike Zheng Shou$^1$\textsuperscript{\Letter}
        }
        \vspace{1.5mm}\\\vspace{-2mm}
	$^1$Show Lab, National University of Singapore\quad
	$^2$Microsoft \\
}
\maketitle

\begin{abstract}
Building Graphical User Interface (GUI) assistants holds significant promise for enhancing human workflow productivity. 
While most agents are language-based, relying on closed-source API with text-rich meta-information (\eg, HTML or accessibility tree), they show limitations in perceiving UI visuals as humans do, highlighting the need for GUI visual agents. 
In this work, we develop a vision-language-action model in digital world, namely \textup{\our}, which features the following innovations: 
\textbf{(i) UI-Guided Visual Token Selection} to reduce computational costs by formulating screenshots as an UI connected graph, adaptively identifying their redundant relationship and serve as the criteria for token selection during self-attention blocks;
\textbf{(ii) Interleaved Vision-Language-Action Streaming} that flexibly unifies diverse needs within GUI tasks, enabling effective management of visual-action history in navigation or pairing multi-turn query-action sequences per screenshot to enhance training efficiency; 
\textbf{(iii) Small-scale High-quality GUI Instruction-following Datasets} by careful data curation and employing a resampling strategy to address significant data type imbalances. 
With above components, \our, a lightweight 2B model using 256K data, achieves a strong 75.1\% accuracy in zero-shot screenshot grounding. Its UI-guided token selection further reduces 33\% of redundant visual tokens during training and speeds up the performance by 1.4×. 
Navigation experiments across web~\cite{mind2web}, mobile~\cite{aitw}, and online~\cite{miniwob++} environments further underscore the effectiveness and potential of our model in advancing GUI visual agents.
The models are available at \textcolor{citecolor}{\url{https://github.com/showlab/ShowUI}}.
\end{abstract}

\begin{figure}[!t]
    \centering
    \includegraphics[width=1.0\linewidth]{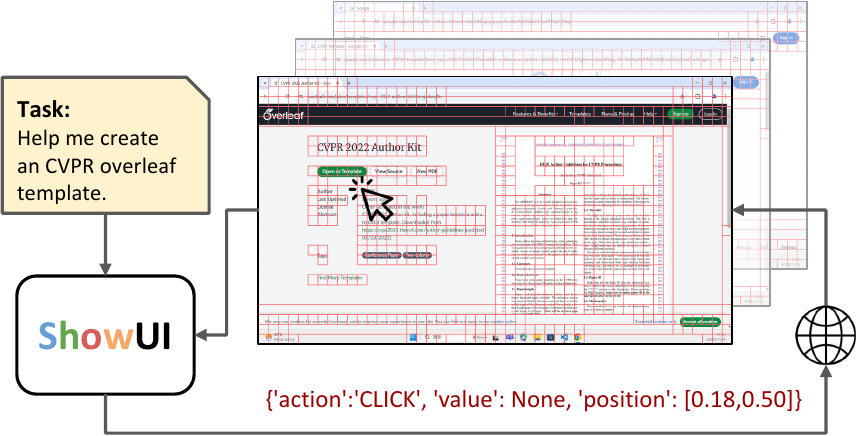}
    \caption{\our~is a Vision-Language-Action model for GUI Automation. 
Given an environment screenshot, \our~efficiently processes it using UI-guided token selection for visual modeling and outputs an interaction action within the loop.
}
    \vspace{1em}    
    \begin{minipage}{0.43\linewidth}
        \centering
        \includegraphics[width=\textwidth]{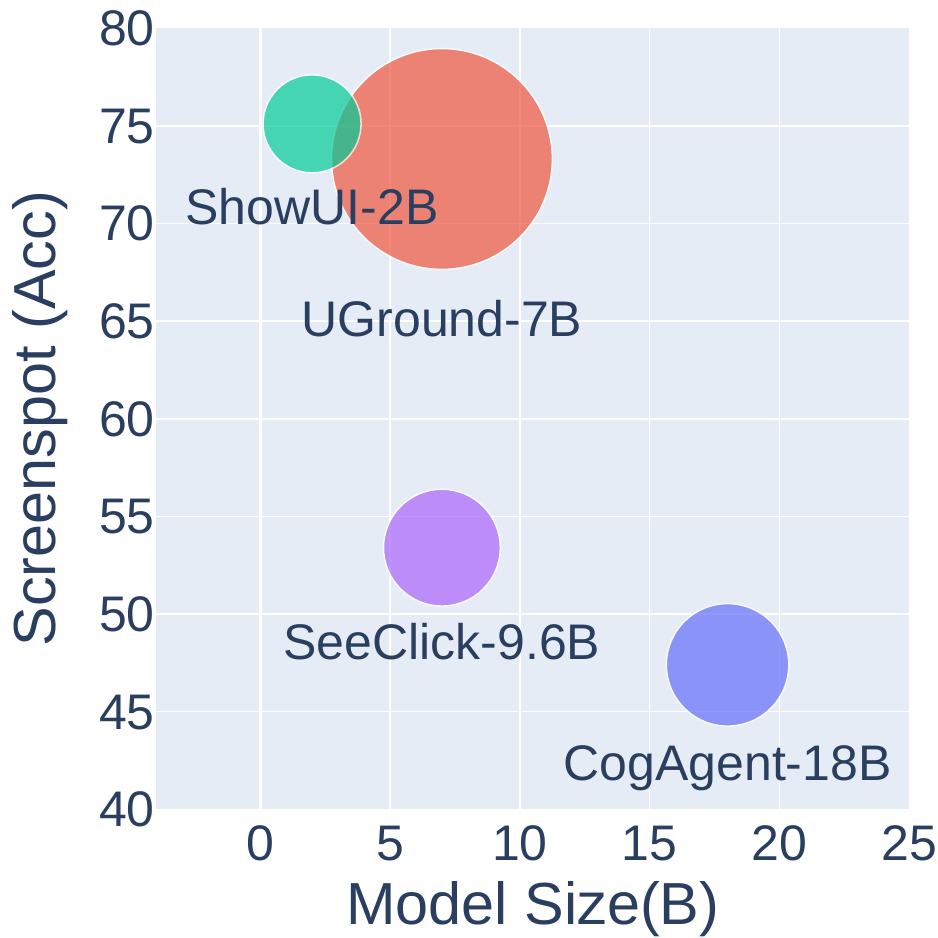}
        \label{point}
    \end{minipage}
    \hfill
    \begin{minipage}{0.55\linewidth}
        \centering
        \includegraphics[width=\textwidth]{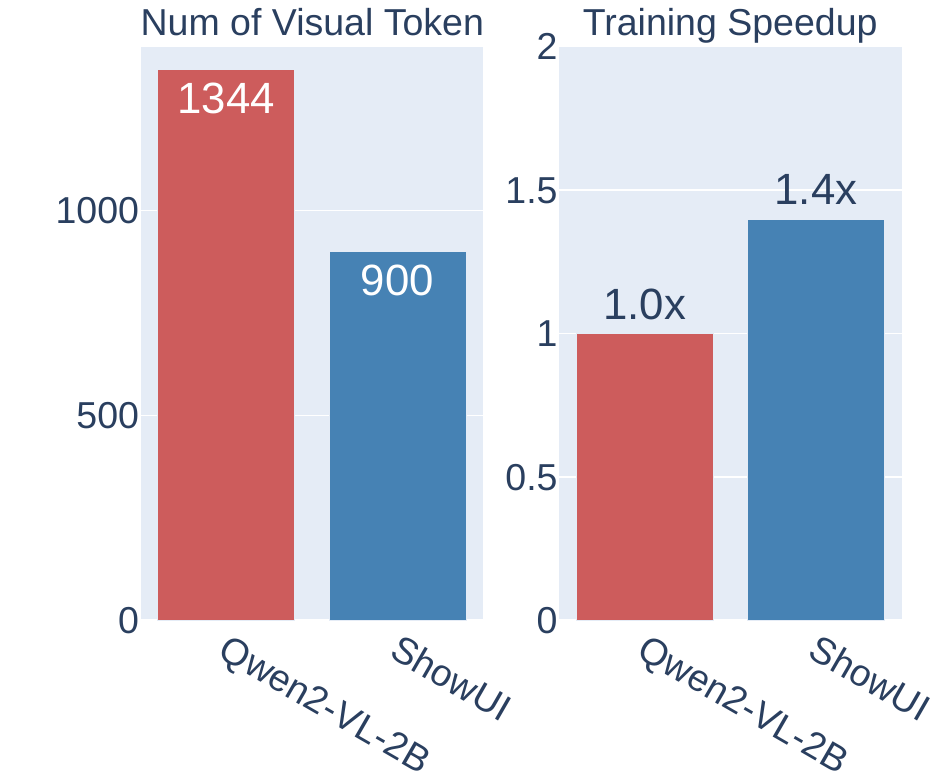}
        \label{bar}
    \end{minipage}   
    \label{fig:teaser}
\vspace{-1.5em}
\caption{
\textbf{Left:} Zero-shot Screenspot grounding comparison between ShowUI and other GUI visual models in terms of \textbf{model size and training scale (area)}; \our~reaching state-of-the-art accuracy as well as the most lightweight model (2B) with a smaller training dataset (256K).
\textbf{Right:} Building upon \base~\cite{qwen2vl}, our UI-guided visual token selection reduces visual token redundancy by 33\% during training, achieving a $1.4\times$ speedup.
}
\vspace{-2em.}
\end{figure}
\section{Introduction}

Graphical User Interfaces (GUIs) are central to how individuals engage with the digital world, serving as virtual embodied interface for a range of daily activities. 
Meanwhile, Large Language Models (LLMs) \cite{gpt4}, with their ability to comprehend complex language instructions and seamlessly integrate tools, have shown significant potential in performing complex tasks through building agents \cite{copilot, webagentplan, webarena, multimodalweb}. 
This progress inspires the development of intelligent GUI agents that can significantly streamline human workflows based on user intentions.


Early efforts in GUI automation have primarily focused on developing \textit{language agents}~\cite{mind2web,gpt4vsom,zheng2024seeact} that rely on closed-source, API-based LLMs like GPT-4~\cite{gpt4}. These agents leverage text-rich metadata like HTML inputs or accessibility trees to perform navigation and other tasks.
However, the text-only approach is limited in real-world applications,
where users typically interact with user interfaces visually—through screenshots—without access to the underlying structural oracle information. This limitation underscores the need for developing GUI \textit{visual agents} that can perceive and interact with UIs as humans do, such as assisting slide creation~\cite{videogui}.

However, GUI visual perception presents unique challenges compared to natural image processing, requiring specialized skills such as UI element grounding or action execution rather than the conversational abilities typical of multi-modal chatbots. 
Recognizing this gap, researchers have begun training vision-language models to acquire these new abilities. 
For instance, studies like \cite{cogagent, seeclick, uground} utilize web screenshot datasets to enhance large multi-modal models' element-grounding capabilities. Meanwhile, works like \cite{guicourse, guiodyssey} address navigation tasks by instruction tuning models for multi-step navigation.

Despite these advancements, training multi-modal models for GUI visual agents continues to face significant challenges related to modeling and training: 
\textbf{({a}) Expensive Visual Modeling:} UI screenshots are typically high-resolution (\eg~2K), resulting in lengthy token sequences that pose issues with long-context processing. Most existing models are not optimized for such high-resolution data, leading to inefficiencies and high computational costs. 
\textbf{(b) Managing Interleaved Vision-Language-Action:} actions differ from language modalities and may vary across devices (\eg~`Return' on web interfaces versus `Press home' on mobile devices) and adapt to different parameter settings (\eg~`Scroll' actions have two directions on the web but four on mobile platforms), how to effectively model action is unclear.
Additionally, it's essential to model actions alongside visual and query data. For example, navigation processes generate a history of screenshot and action steps, creating a complex, interleaved vision-language-action that models must effectively interpret and manage.
\textbf{({c}) Diverse Training Data:} with vast amounts of GUI data across different devices such as Web and Mobile, accompanied by diverse purpose annotations including element grounding and navigation, it remains unclear how to effectively select a high-quality training corpus for developing robust GUI visual models.
These critical challenges have been under-explored but are essential for the development of effective visual models for GUI agents. 

In this work, we develop a vision-language model for GUI visual agents, aiming to address and resolve the aforementioned challenges, with the following key contributions:

\noindent\textbf{(\textit{i}) UI-Guided Visual Token Selection:} 
We recognize the uniqueness of UI screenshots (\ie, redundancy mixed with essential details) and develop a UI-friendly visual token selection approach. In RGB space, we represent each patch as a node and identify connected components to model redundancy across patches. This relationship guides the self-attention blocks within visual encoders or language models for token selection, effectively reducing computation.

\noindent\textbf{(\textit{ii}) Interleaved Vision-Language-Action Streaming:}
We analyze the diversity of GUI actions, structuring them in JSON format and documenting their action space to assist the model in action interpretation. 
Additionally, we recognize the need for interleaved understanding across modalities, such as combining action with visual navigation history and balancing visual token lengths through multi-turn action with text queries to improve training efficiency. 
Our model is formulated as interleaved vision-language-action streaming, unifying the diverse needs in GUI scenarios.

\noindent\textbf{(\textit{iii}) Well-selected Instruction-following Dataset:} 
Instead of utilizing data from all available sources, we conduct an in-depth analysis of each data type’s properties. For example, in web data, visual elements (\ie, button) is more valuable than static text (which accounts for 40\%) as most VLMs~\cite{qwen2vl} possess strong OCR capabilities. 
Additionally, we introduce a small, high-quality instruction-following dataset that achieves strong UI grounding performance. Furthermore, we develop a rebalanced sampling strategy to address the substantial imbalance in UI data, ensuring consistent model performance across different setups.

Building upon aforementioned innovations, we enhance \base\ to create a powerful GUI visual agent, \our. As shown in Fig.\ref{point}, this results in a lightweight 2B model using 256K data, achieving a strong 75.1\% accuracy in zero-shot screenshot grounding. 
\our~also demonstrate competitive navigation ability in Web~\cite{mind2web}, Mobile~\cite{aitw}, and Online~\cite{miniwob++} environments. 
Comprehensive ablation studies (Fig.~\ref{bar}) demonstrate the effectiveness of our UI-guided token selection approach, reducing redundant visual tokens by 33\% and accelerating training by 1.4×. 
Moreover, we conclude with many discussions on current performance gaps and future directions.


\begin{figure*}[!t]
  \centering
  \includegraphics[width=\textwidth]{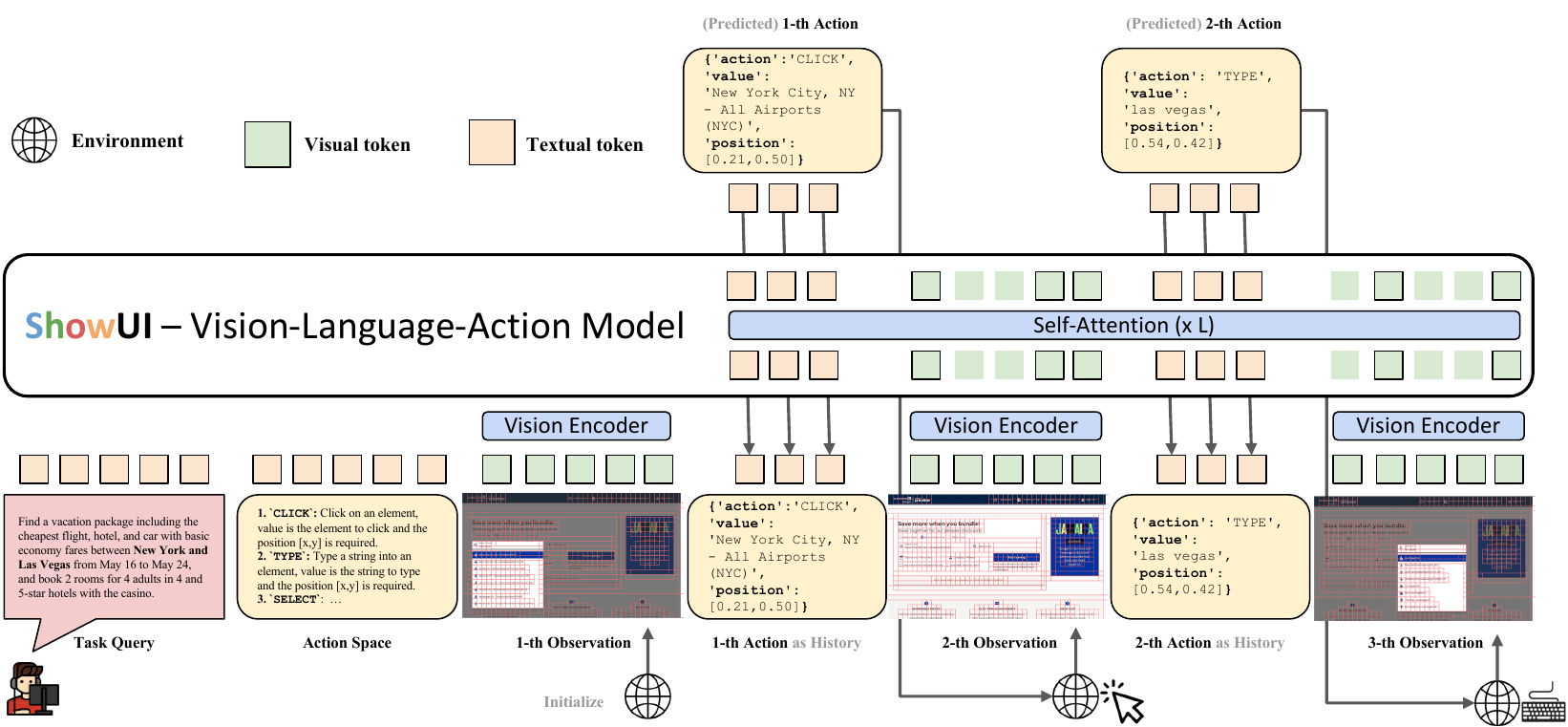} %
\caption{\textbf{Illustration of \our.}
Given a user task query, a pre-defined action space, and an initial screenshot as observation, \our~proceeds by executing the next action, updating the screenshot, and continuing in this cycle.
Notably, \our~features the following key innovation designs:
\textbf{(i) UI-Guided Visual Token Selection}, which processes the input screenshot to build UI patch-wise connected graph. During training, a random subset of tokens is selected within each component for efficient visual modeling (Sec.~\ref{sec:uiguided}).
\textbf{(ii) Interleaved Vision-Language-Action Streaming} to effectively handle past screenshots and actions, improving navigation performance. (Sec.~\ref{sec:stream}).
}
\label{fig:vla}
\end{figure*}

\section{\SHOW}
\our, as outlined in Fig.~\ref{fig:vla}, is built on top of the vision-language model \base~\cite{qwen2vl}, incorporating the following key components optimized for GUI tasks:
\textbf{(i)} a novel UI-guided visual token selection strategy for efficient visual modeling, 
\textbf{(ii)} an interleaved vision-language-action streaming setup to flexibly unify different needs by GUI tasks and enhance training effectiveness, 
\textbf{(iii)} a training data recipe, crafted through a detailed analysis of individual GUI data types, 
which enables ShowUI training on a smaller, high-quality corpus. 
In the following sections, we introduce each component in detail.

\subsection{UI-Guided Visual Tokens Selection}\label{sec:uiguided}
\begin{figure*}[!t]
    \centering
    \begin{subfigure}[b]{0.59\textwidth}
        \centering
      \includegraphics[width=1\textwidth]{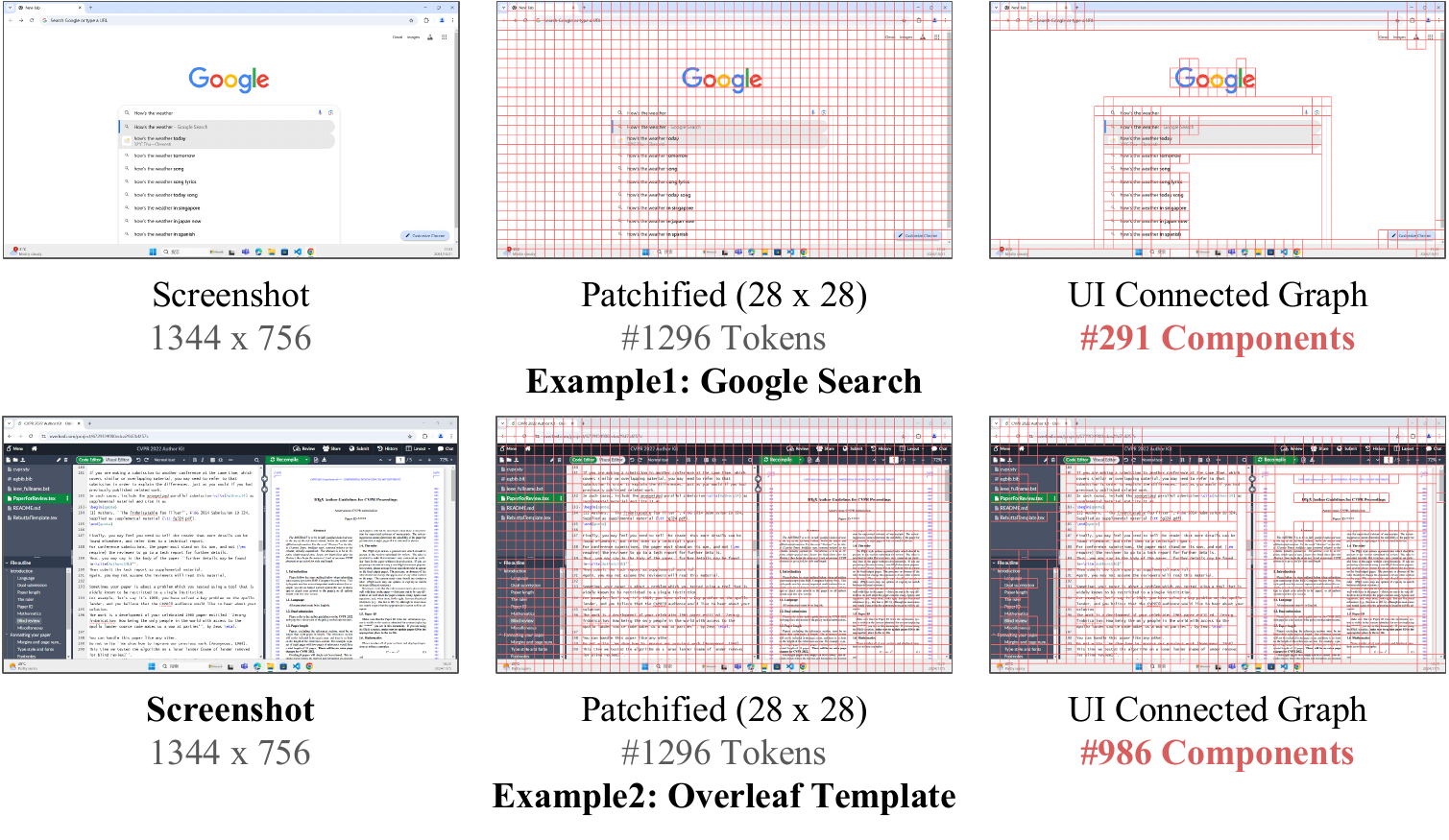} %
        \caption{UI Connected Graph adaptively assigns connected components based on the informativeness of screenshots.}
        \label{fig:uigraph:left}
    \end{subfigure}
    \hfill
    \begin{subfigure}[b]{0.39\textwidth}
        \centering
      \includegraphics[width=0.9\textwidth]{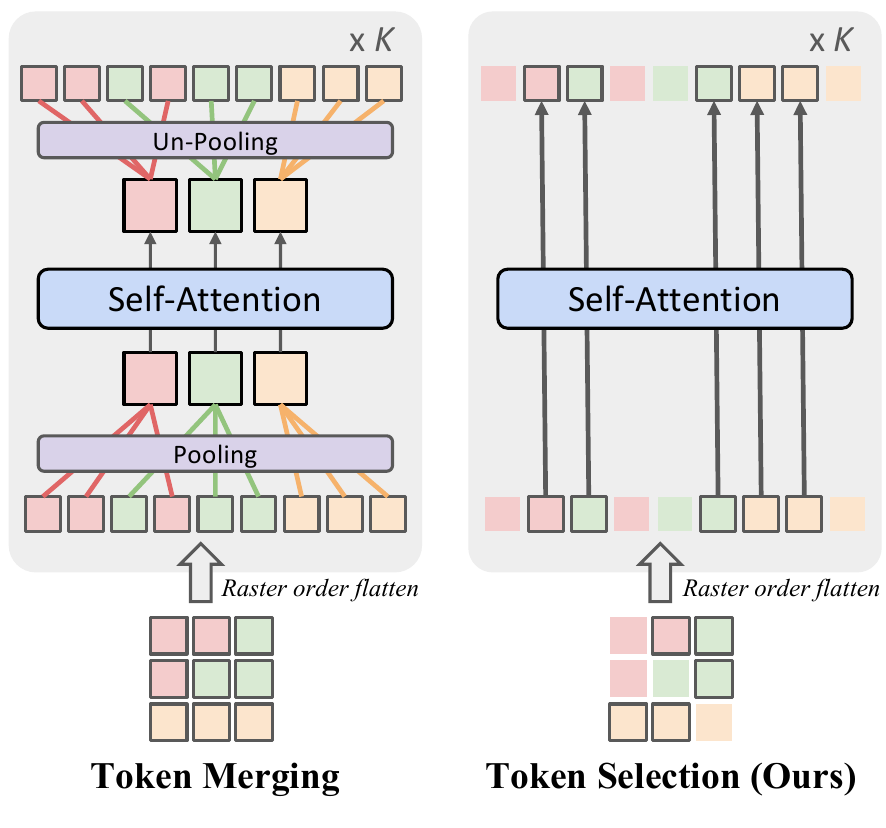} %
        \caption{Two representative token compression methods, where patches of the same color indicate the same component and are redundant to each other.}
        \label{fig:uigraph:right}
    \end{subfigure}
 \vspace{1em}
\caption{
\textbf{Illustration of UI-guided Visual Tokens Selection.}
\textbf{Left:} Starting from an original screenshot (left) with a 28x28 patch grid (middle), resulting in 1296 tokens, the UI Connected Graph adaptively organizes the image into disjoint regions based on RGB values, allowing patches within the same region to be treated as redundant.
\textbf{Right:} Comparison of two methods leveraging UI Connected Graph in visual token modeling: Token merging pools all tokens within one component, which loses individual positional information, while token selection, randomly sample part of tokens with each component, still retains their original position relationship.
}
\label{fig:main}
\end{figure*}

High-resolution screenshots can result in a large number of visual tokens after standard patching. 
As demonstrated in Fig.\ref{fig:uigraph:left}, a $1344\times 756$ resolution on a PC yields approximately $5184$ raw tokens with a $14\times14$ patching, after a $2\times2$~\cite{qwen2vl} merging, still results in $1296$ tokens, creating a computational challenge within the self-attention module.


\noindent\textbf{What differentiates UI from natural vision?} 
Unlike natural images, which captures real-world complexities and unpredictable patterns thus rich in semantic, textures, UI screenshots are inherently structured, with clear layouts and consistent color schemes optimized for readability and usability. 
This difference means that UI images often contain {redundant} empty spaces or simple backgrounds that do not carry essential information, aling for optimization or pruning. Moreover, small but functionally important elements, like icons or text, demand higher {salience} due to their role in interactivity and clarity. 

Thus, it is necessary to have a strategy that can \textit{differentiate between redundant and essential visual elements}, enabling effectively pruning of irrelevant visual tokens without compromising usability. 
We found that the \textbf{RGB space} can serve as a useful guideline for this purpose as pattern variants, text fonts can be easily identify by its RGB values.

\begin{algorithm}[!b]
\small
\caption{Find Connected Components on UI-Graph}
\begin{algorithmic}[1]
    \State \textbf{Input:} Screenshot of size \( H \times W \), patch size \( c \), threshold \( \delta \)
    \State \textbf{Output:} Assignment map between patch and connected components.
    
    \State Divide the image into \( G_h \times G_w \) patches, each patch is a node, where \( G_h = \frac{H}{c} \) and \( G_w = \frac{W}{c} \)
    \State Initialize Union-Find structure \texttt{UF} over nodes
    
    \ForAll{node \( (i, j) \)}
        \ForAll{neighbors \( (i', j') \) to the right and below of \( (i, j) \)}
            \If{ \( \| \text{RGB}\left(i, j\right) - \text{RGB}\left(i', j'\right) \| < \delta \) }
                \State \texttt{UF.union} \(\left((i, j), (i', j')\right)\)
            \EndIf
        \EndFor
    \EndFor
    
    \State \Return Assignment map from \texttt{UF}
\end{algorithmic}
\label{algorithm:uigraph}
\end{algorithm}

\noindent\textbf{Construct a UI Connected Graph.}
After dividing the screenshot into regular patches, we observed many neighboring patches share exactly the same RGB values and are thus redundant. To leverage this, we represent each patch as a node in a graph. If neighboring patches have the same RGB values, we connect their corresponding nodes, forming \textbf{connected components}. 
This allows us to group and simplify redundant areas while preserving essential visual elements identified by their unique RGB patterns.
Visually identical patches can be easily detected by setting a small threshold on the difference between their patch tensors. 
Based on this insight, we use the Union-Find method to identify connected components in this UI connected graph, as described in Algorithm~\ref{algorithm:uigraph}.
This algorithm produces a graph with $K$ connected component, where $K$ is typically smaller than the original number of patches $G_h \times G_w$. Based on the assignment of each node to its component, we can model the redundancy relationships among patches.

As illustrated in Fig.~\ref{fig:uigraph:left}, this method can effectively balances component number based on their visual informative adaptively, using less component (more redundant patches) in google search page with sparser areas ($1296\rightarrow 291$), while assigning more components (more independent patches) in text-rich overleaf screenshots ($1296\rightarrow 986$).
In Fig.\ref{supp:fig:uigraph}, we display how our method constructs the \textbf{UI-connected graph} across different devices.
Given identical resolution screenshots with the same initial visual patch tokens (\eg, 1272), our method adaptively constructs connected components based on the informativeness of the screenshots.



\begin{figure*}[htbp]
    \centering
    \begin{subfigure}{0.24\textwidth}
        \centering
        \includegraphics[width=\textwidth]{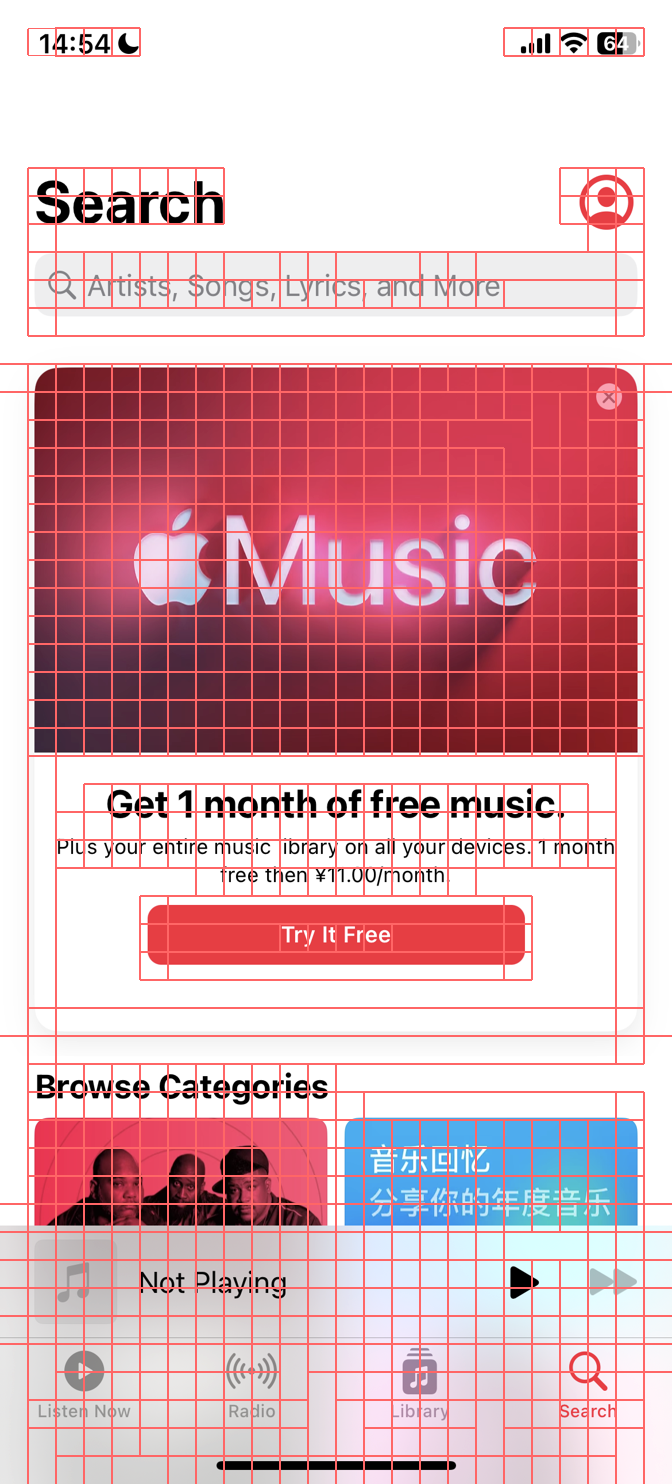}
        \caption{1272 tokens $\rightarrow$ 781 components}
    \end{subfigure}
    \hfill
    \begin{subfigure}{0.24\textwidth}
        \centering
        \includegraphics[width=\textwidth]{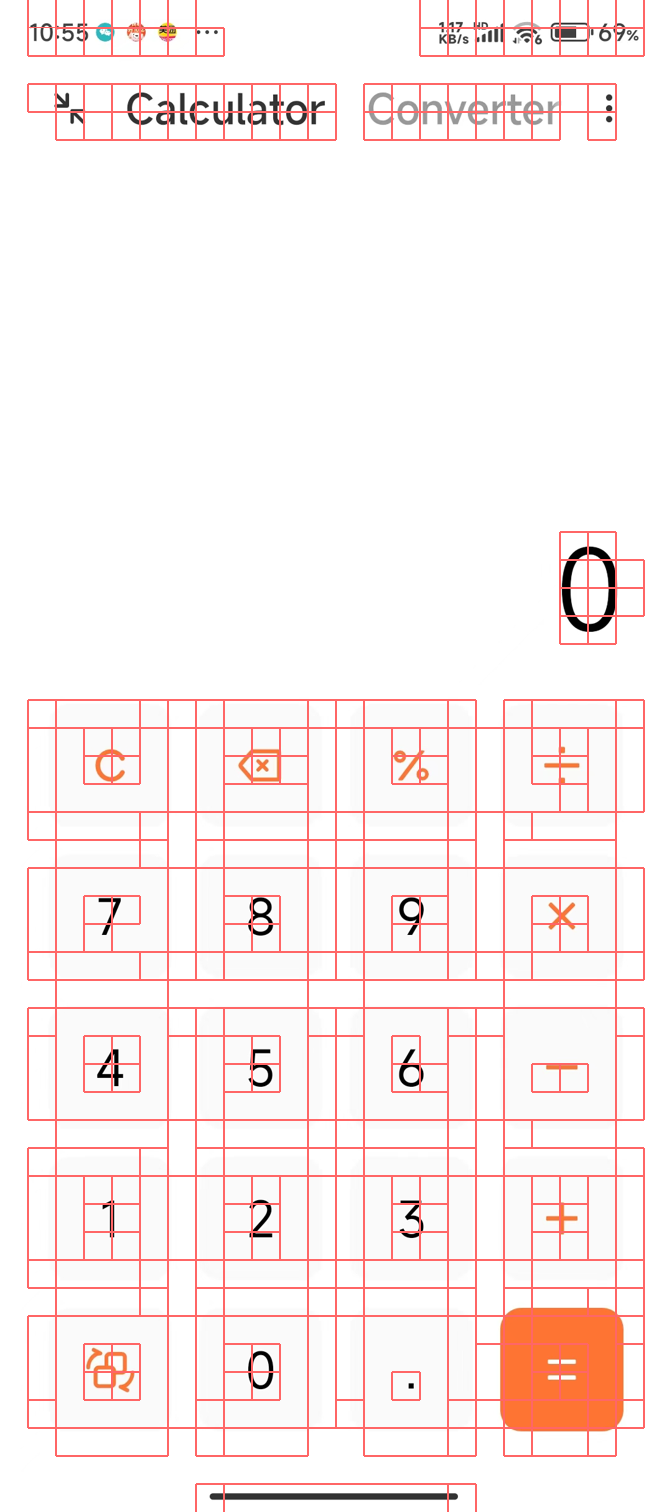}
        \caption{1272 tokens $\rightarrow$ 359 components}
    \end{subfigure}
    \hfill
    \begin{subfigure}{0.24\textwidth}
        \centering
        \includegraphics[width=\textwidth]{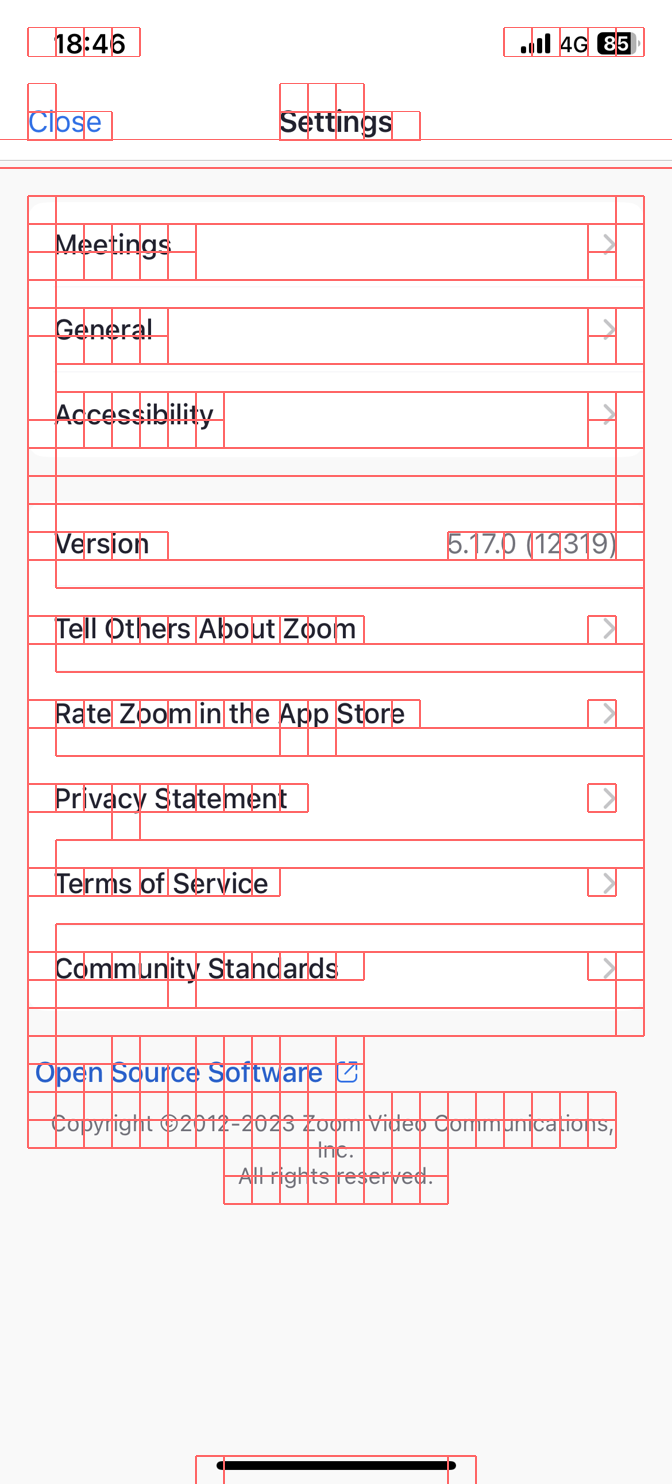}
        \caption{1272 tokens $\rightarrow$ 265 components}
    \end{subfigure}
    \hfill
    \begin{subfigure}{0.24\textwidth}
        \centering
        \includegraphics[width=\textwidth]{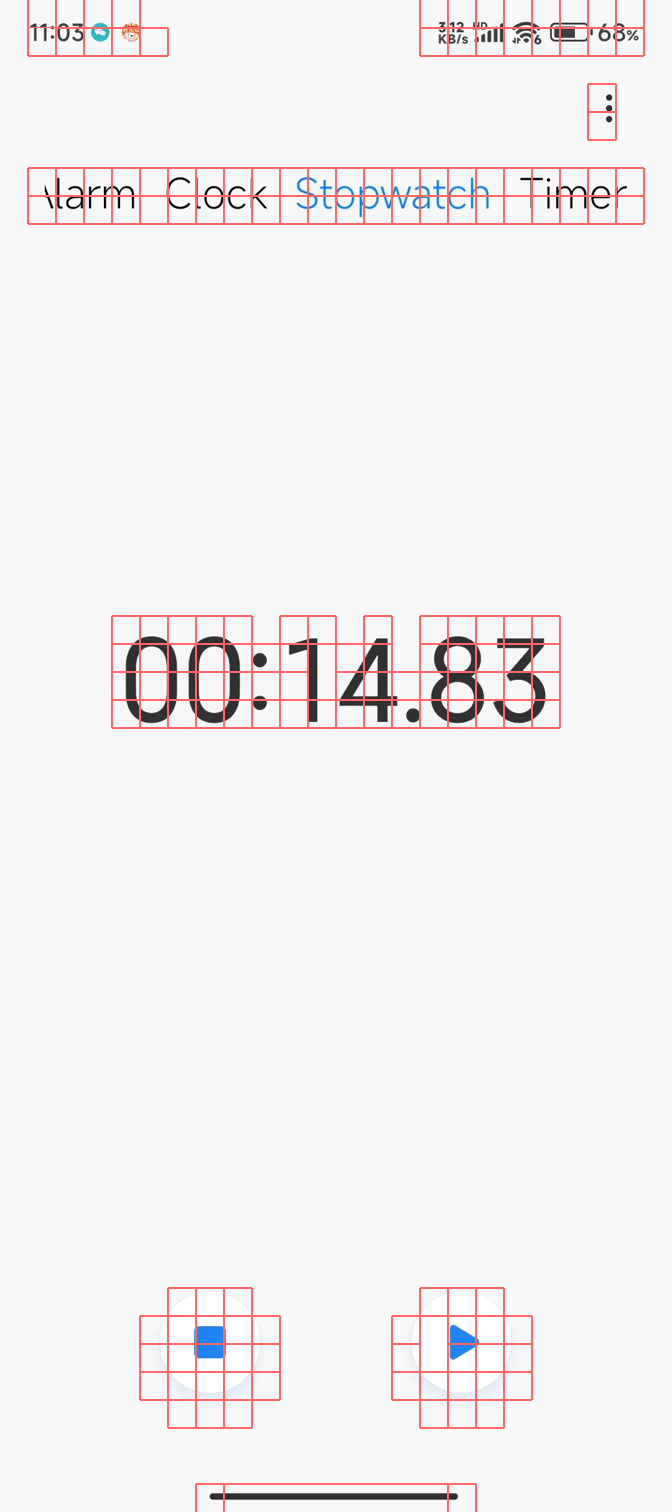}
        \caption{1272 tokens $\rightarrow$ 175 components}
    \end{subfigure}
    
    \vspace{2em} 

    \begin{subfigure}{0.49\textwidth}
        \centering
        \includegraphics[width=\textwidth]{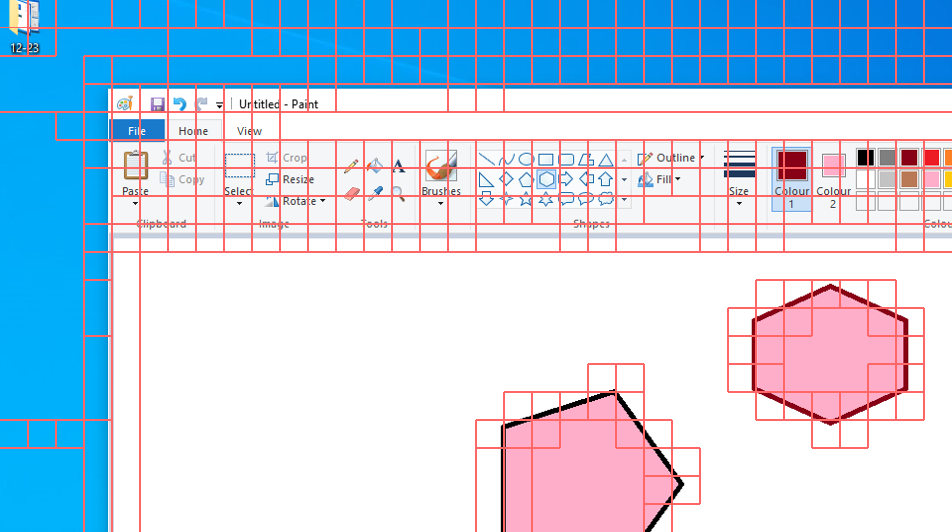}
        \caption{646 tokens $\rightarrow$ 281 components}
    \end{subfigure}
    \hfill    
    \begin{subfigure}{0.49\textwidth}
        \centering
        \includegraphics[width=\textwidth]{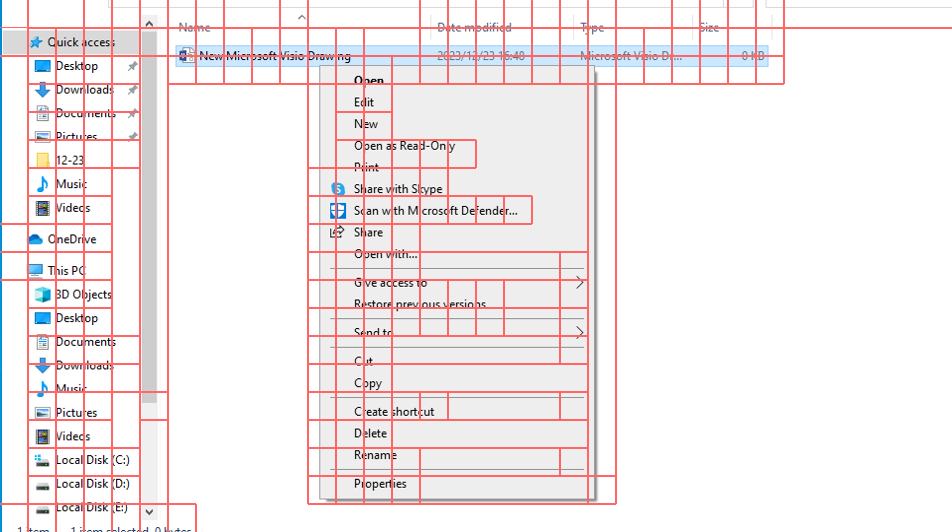}
        \caption{646 tokens $\rightarrow$ 230 components}
    \end{subfigure}
    
    \vspace{2em} 
    \begin{subfigure}{0.49\textwidth}
        \centering
        \includegraphics[width=\textwidth]{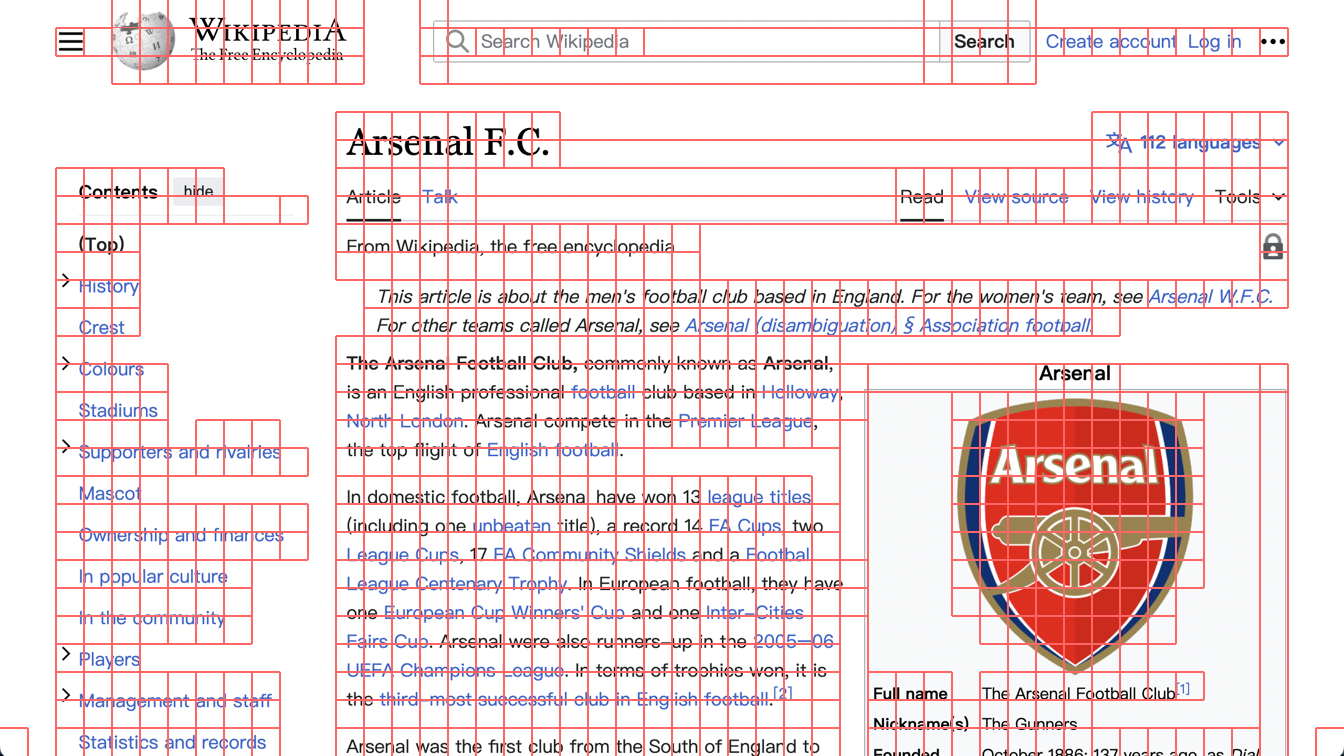}
        \caption{1296 tokens $\rightarrow$ 740 components}
    \end{subfigure}
    \hfill
    \begin{subfigure}{0.49\textwidth}
        \centering
        \includegraphics[width=\textwidth]{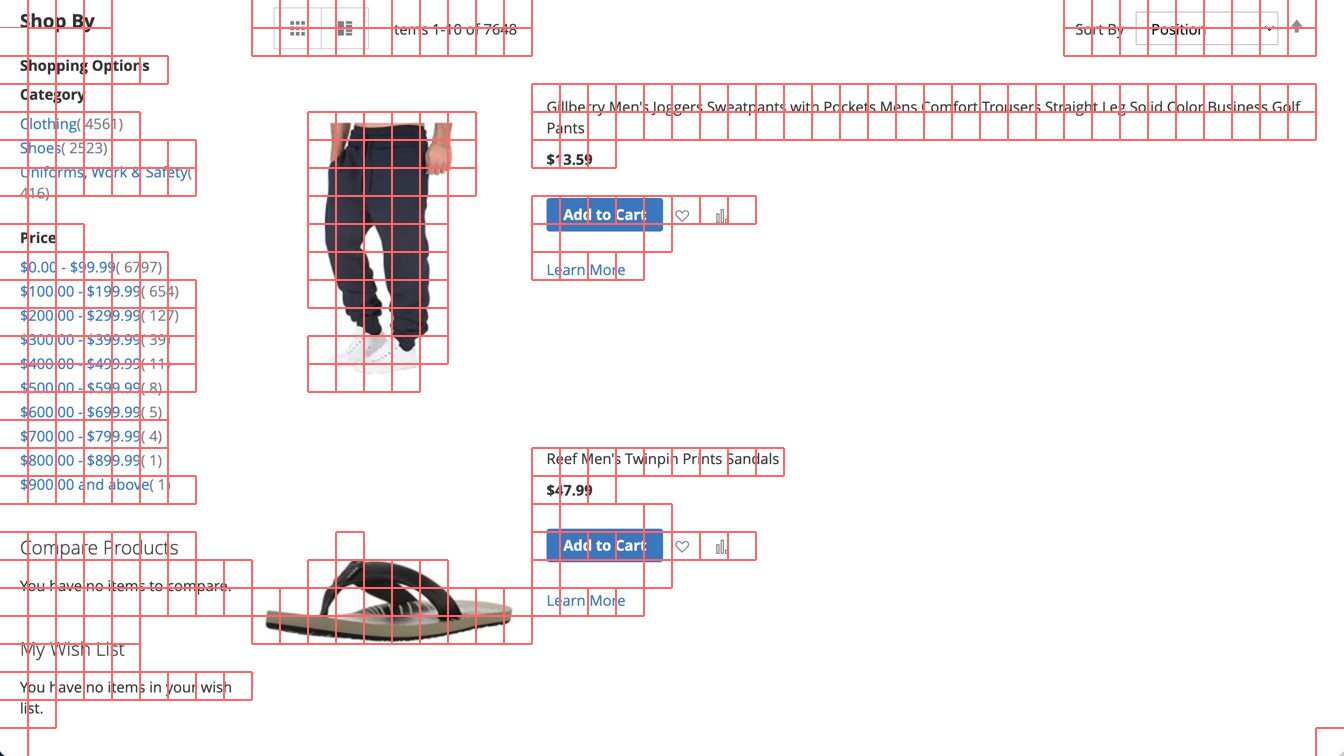}
        \caption{1296 tokens $\rightarrow$ 369 components}
    \end{subfigure}
    
    \vspace{2em} 

    \caption{Illustration of our method constructs the \textbf{UI-connected graph} based on the informativeness of screenshots.
    (a–d) Mobile; (e–f) PC; (g–h) Web.}
    \label{supp:fig:uigraph}
\end{figure*}

\noindent\textbf{Token Merging \textit{v.s.} Token Selection.}
Next, we explore how leveraging this UI connected graph can improve model efficiency. 
We find that the primary computational bottleneck in existing vision-language models lies in the long sequences processed by cascaded self-attention layers, impacting both language models and visual encoders.

A straightforward approach is to apply token merging methods~\cite{tokenmerge,chatuniv}, which represent all patches within a component into a single token, as shown in the Fig.\ref{fig:uigraph:right} left half.
In practice, however, we found that this approach disrupts positional relationships, as the original positional information in the pooled token sequence is inevitably lost, which is essential for accurate UI element grounding.


To enable token compression within self-attention without losing positional information, we draw inspiration from Mixture-of-Depth~\cite{mod}, which sparsely samples tokens via a routing mechanism. In our case, the UI connected graph provides an effective routing criterion. Tokens within the same component can be considered redundant, so we randomly skip a portion of tokens within each component during training, leaving single-patch components unaffected to preserve essential elements. 
For the selected tokens, we \textit{retain} their original position embeddings, ensuring that self-attention operates on the original positional relationships, even with a less token sequence.

Notably, this token selection method introduces no additional learnable parameters. 
Therefore, we apply random token selection at a set ratio during training, while at inference, the method offers flexibility to use either with or without token selection, as both options maintain consistent positional relationships within the full token sequence.

\subsection{Interleaved VLA Streaming}\label{sec:stream}
In this section, we aim to address how to formulate actions and their relationships with other modalities (\ie, visual and textual queries).

\noindent\textbf{What differentiates Action from natural text?}
The core functionality of a GUI model is navigation, conditioned on a text query and requiring the model to jointly predict: the correct action type (\eg, \texttt{[CLICK]} or \texttt{[TYPE]}), with corresponding action parameters (\eg, coordinates for \texttt{[CLICK]} or a string for \texttt{[TYPE]}).
A major challenge in navigation arises from the action variants across different devices, such as:
\textit{(i)} Device-specific actions (\eg~\texttt{[CLICK]} is unavailable on mobile, whereas \texttt{[PRESS HOME]} does not exist on the web).
\textit{(ii)} Same action with different parameters (\eg~\texttt{[SCROLL]} has two directions---up and down---on the web, but four directions on mobile).
\textit{(iii)} Novel actions at test time that were not encountered during training.

\begin{figure}[!t]
    \centering
    \includegraphics[width=1.0\linewidth]{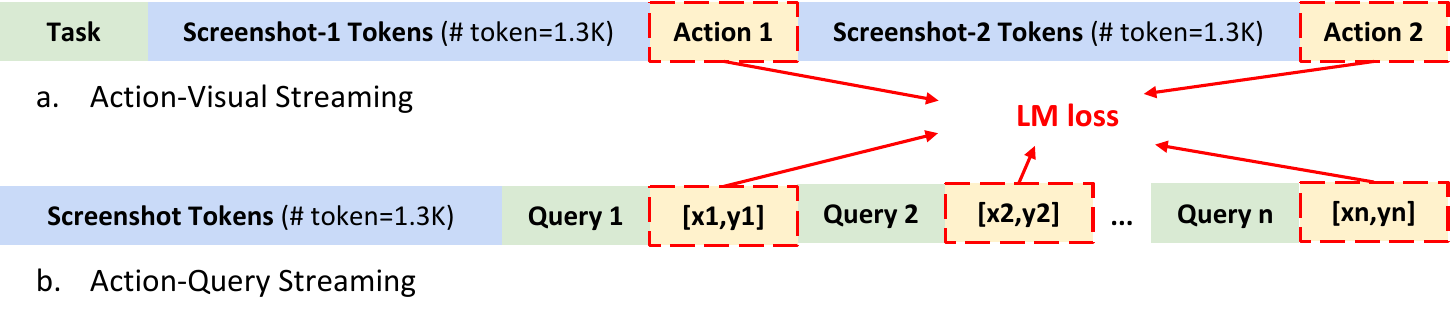}
    \caption{\textbf{Illustration of Interleaved Vision-Text-Action Streaming.} 
    The visual tokens in screenshots are significantly longer (\eg, 1.3K) compared to queries or actions (\eg, fewer than 10). 
    Thus, we introduce two modes: \textbf{(a)} Action-Visual Streaming for UI navigation and \textbf{(b)} Action-Query Streaming for UI grounding. These modes extend the concept of multi-turn dialogue and enable more efficient utilization of training data.
    }
    \label{fig:streaming}
\end{figure}

To manage these variations within our model, we first structure each action in a \textbf{JSON} format (\ie, 
{\{\texttt{‘action’: ‘action\_type’, ‘value’: ‘element’, ‘position’: [x,y]}\}}), where \texttt{[x,y]} represents relative coordinates on 0-1. This allows us to standardize actions from diverse devices into a unified format.
Secondly, we provide the model with a ‘\textbf{README}’ in the system prompt, documenting each action's usage within the action space (\eg, ‘CLICK’: Click on an element, value is not applicable and the position [x,y] is required.)
This setup encourages the model to interpret the action space document rather than memorizing fixed actions, enabling it to execute actions at test time in a function-calling manner~\cite{xlam}. Next, we discuss the relationship between action and other modalities, as illustrated in Fig.\ref{fig:streaming}.

\noindent\textbf{Action with Visual:}
The GUI navigation process typically involves multi-step trajectories, requiring the model to 
recognize current steps and determine the next action in context.
This introduces the challenge of managing both past actions and past observations (screenshots): actions indicate what has been done but lack visual context, while screenshots capture the visual state but omit actions taken. 
To ensure complete history information, we formulate an interleaved vision-action stream, as shown in Fig.\ref{fig:vla}, which captures both visual and action information sequentially. After the $i$-th action, the resulting $(i+1)$-th screenshot enters the queue following the previously executed $i$-th action, prompting the model to generate the $(i+1)$-th action.

In practice, we can optionally mask parts of the visual history depending on the application. For instance, in Mobile~\cite{aitw}, where cross-software screenshot changes occur, it is essential to retain screenshots to track visual appearances. In contrast, for Web~\cite{mind2web}, where screenshots generally remain stable on a static webpage across a series of actions, masking may be preferable for efficiency.

\noindent\textbf{Action with Textual query:}
In one-step navigation or element grounding, we might encounter one screenshot with multiple parallel actions~\cite{omniact} or multiple elements~\cite{seeclick}, where screenshots tend to be high-resolution, resulting in long token sequences (\eg, 1-2K tokens). Meanwhile, queries like UI element names and actions (coordinates) are typically much shorter (often fewer than 10 tokens). 
This discrepancy makes a one-image-per-action approach inefficient. To optimize training data utilization, we adopt a multi-turn dialogue approach, predicting multiple action annotations for each screenshot within a single forward pass.

\subsection{GUI Instructional Tuning}\label{sec:data}

Various GUI datasets are available in community, such as dominant of web data~\cite{webui,seeclick,guicourse}, mobile data~\cite{aitw,guiodyssey}, which may contain element coordinates~\cite{seeclick} or user trajectories~\cite{guiodyssey}.
Rather than aggregating all available data sources, we analyze each dataset type to select representative data.
Our selected data is illustrated in Tab.\ref{table:dataset}. 
Our discussion is mainly on UI grounding data. 
For navigation, we source GUIAct~\cite{guicourse} with mobile and web devices.


\noindent\textbf{(i) Web--visual elements:} The web provides a highly accessible, text-rich source of UI data, easily crawled from HTML~\cite{webui}. Our statistical analysis shows that the 'static text' tag accounts for a significant portion (40\%). 
Given that most VLMs already with strong OCR capabilities~\cite{phi3,qwen2vl}, we focus on collect visually rich elements. 
To this end, we developed a parser and collected 22K screenshots, retaining only visual-related elements such as those tagged with ‘Button’ or ‘Checkbox’. By removing static text

\begin{figure}[!h]
    \centering
    \includegraphics[width=1.0\linewidth]{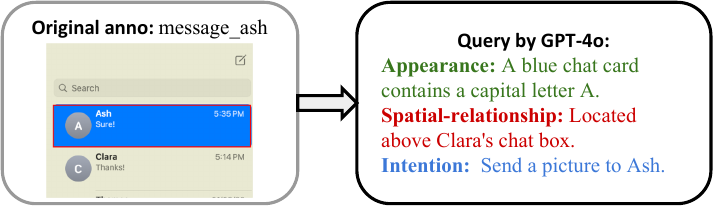}
    \caption{We derive three types of query (appearance, spatial relationship, and intention) from raw annotation, assisted by GPT-4o.}
    \label{fig:gpt4o}
\end{figure}

\noindent\textbf{(ii) Desktop--diverse query:} Desktop data is particularly valuable as it is challenging to collect automatically. 
We identified OmniAct~\cite{omniact}, which includes manual elements from iOS, Windows, and Linux desktops with a small size (2K elements over 100 images). Its element is only labelled by original name such as ‘message\_ash’.
To enrich this dataset and diversity, we employed {reverse engineering} techniques, utilizing ground-truth bounding boxes and its text elements. Then we prompt GPT-4o, with visual prompts highlighting target elements, to derive three types of query: appearance, spatial and intention; which we illustrated in Fig.\ref{fig:gpt4o}. 
This method results new 6K elements. See Supp. for detail prompt and discussion.

\begin{figure*}[htbp]
    \centering
    \begin{subfigure}{0.49\textwidth}
        \centering
        \includegraphics[width=\textwidth]{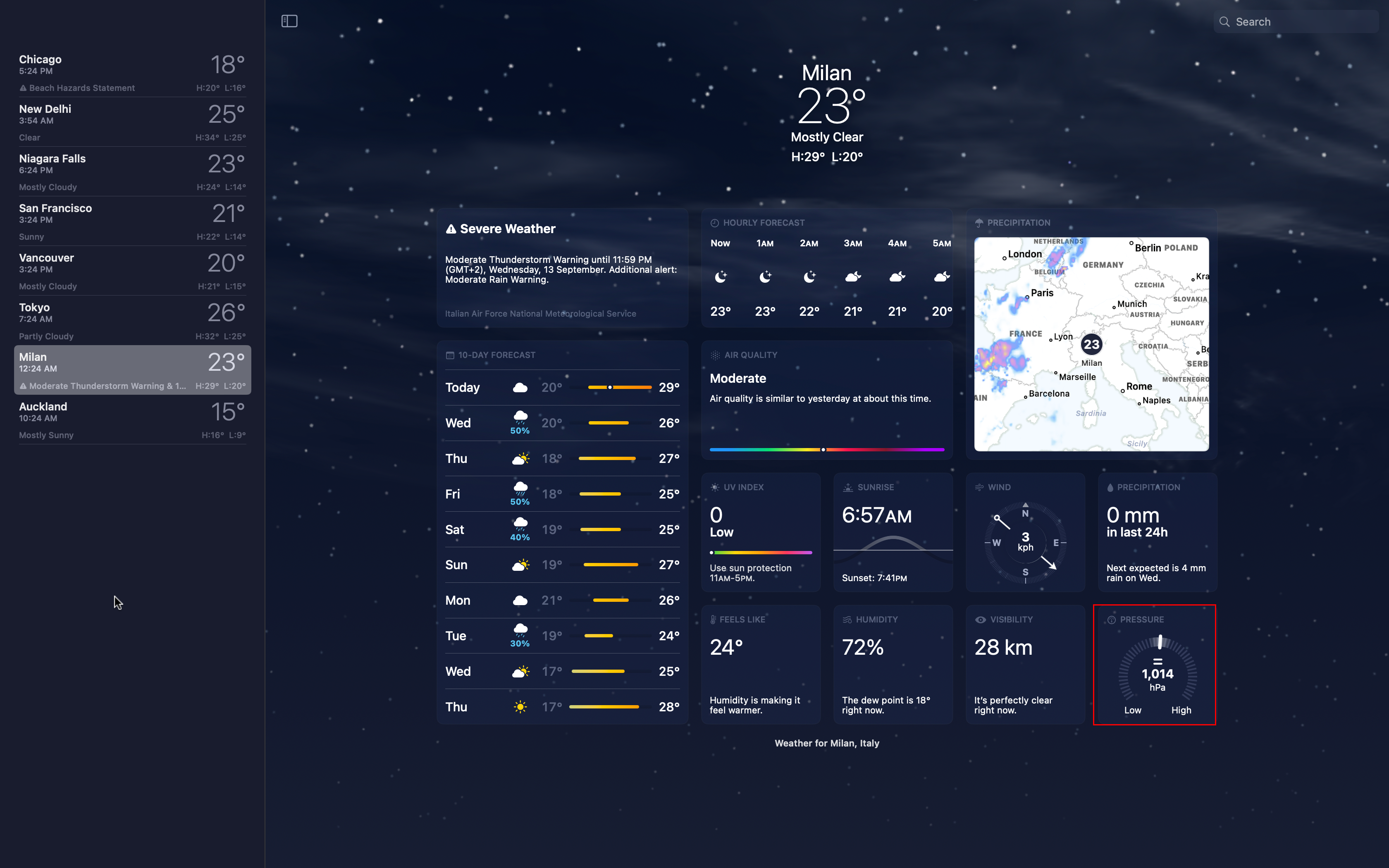}
        \caption{
            Example of Weather. \textbf{Original:} ‘visibility’; 
            \textcolor{darkgreen}{\textbf{Appearance:} “A rectangular box with 28 km in white text.”};
            \textcolor{darkred}{\textbf{Spatial:} “Positioned below ‘WIND’ and next to ‘PRESSURE’.”};
            \textcolor{darkblue}{\textbf{Intention:} “Check current fog or mist conditions.”}}
    \end{subfigure}
    \hfill
    \begin{subfigure}{0.49\textwidth}
        \centering
        \includegraphics[width=\textwidth]{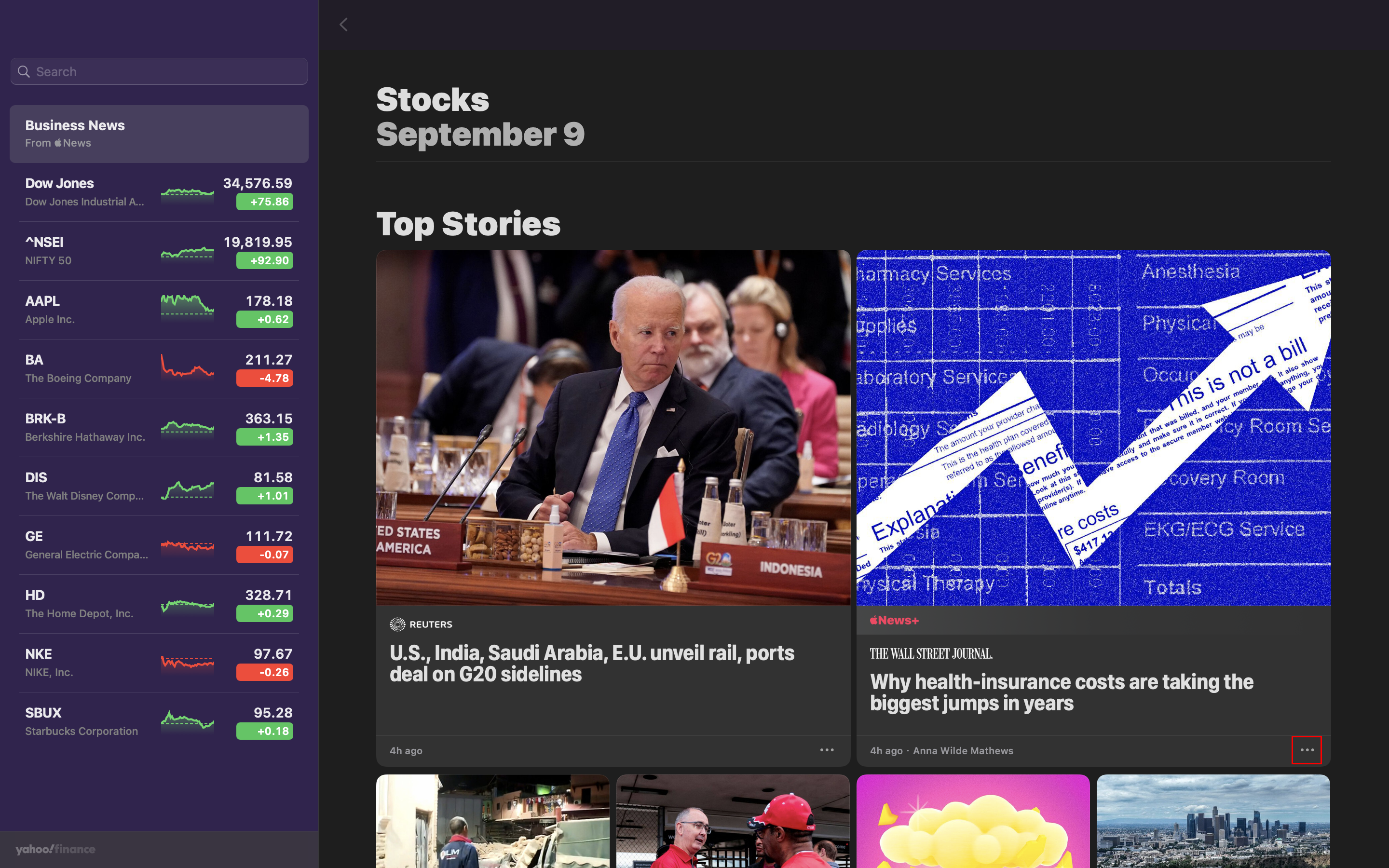}
        \caption{
            Example of Stocks. \textbf{Original:} ‘Share option-health insurance’;
            \textcolor{darkgreen}{\textbf{Appearance:} “Three vertical dots icon on a dark background.”};
            \textcolor{darkred}{\textbf{Spatial:} “Located to the right of the health insurance headline.”};
            \textcolor{darkblue}{\textbf{Intention:} “Click to share the health insurance article.”}}
    \end{subfigure}

    \vspace{2em} 

    \begin{subfigure}{0.49\textwidth}
        \centering
        \includegraphics[width=\textwidth]{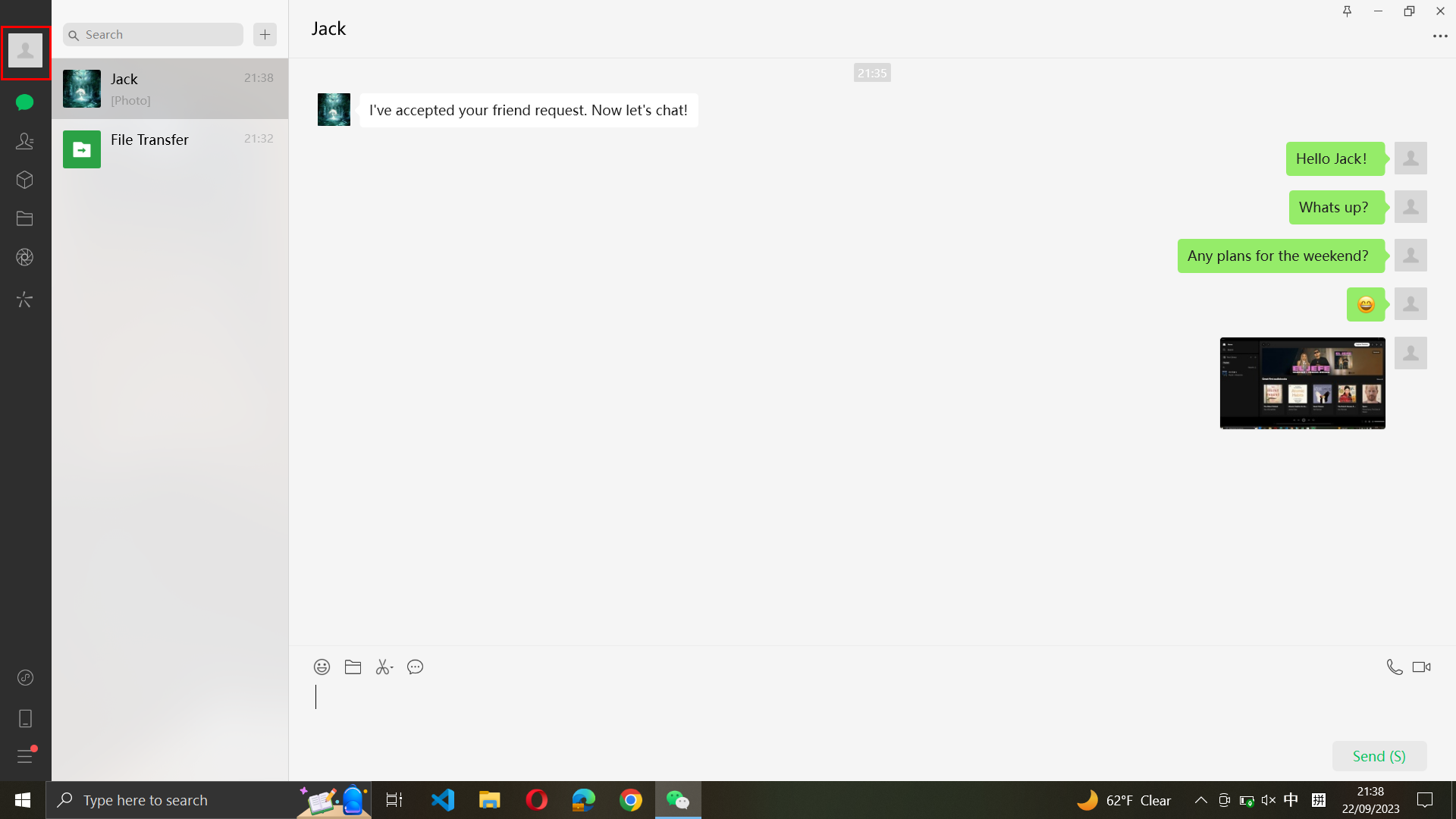}
        \caption{
            Example of WeChat. \textbf{Original:} ‘expand\_profile’;
            \textcolor{darkgreen}{\textbf{Appearance:} “A rounded gray button with a person icon.”};
            \textcolor{darkred}{\textbf{Spatial:} “Located at the top-left corner of the chat pane.”};
            \textcolor{darkblue}{\textbf{Intention:} “Expand the contact's profile view.”}}
    \end{subfigure}
    \hfill
    \begin{subfigure}{0.49\textwidth}
        \centering
        \includegraphics[width=\textwidth]{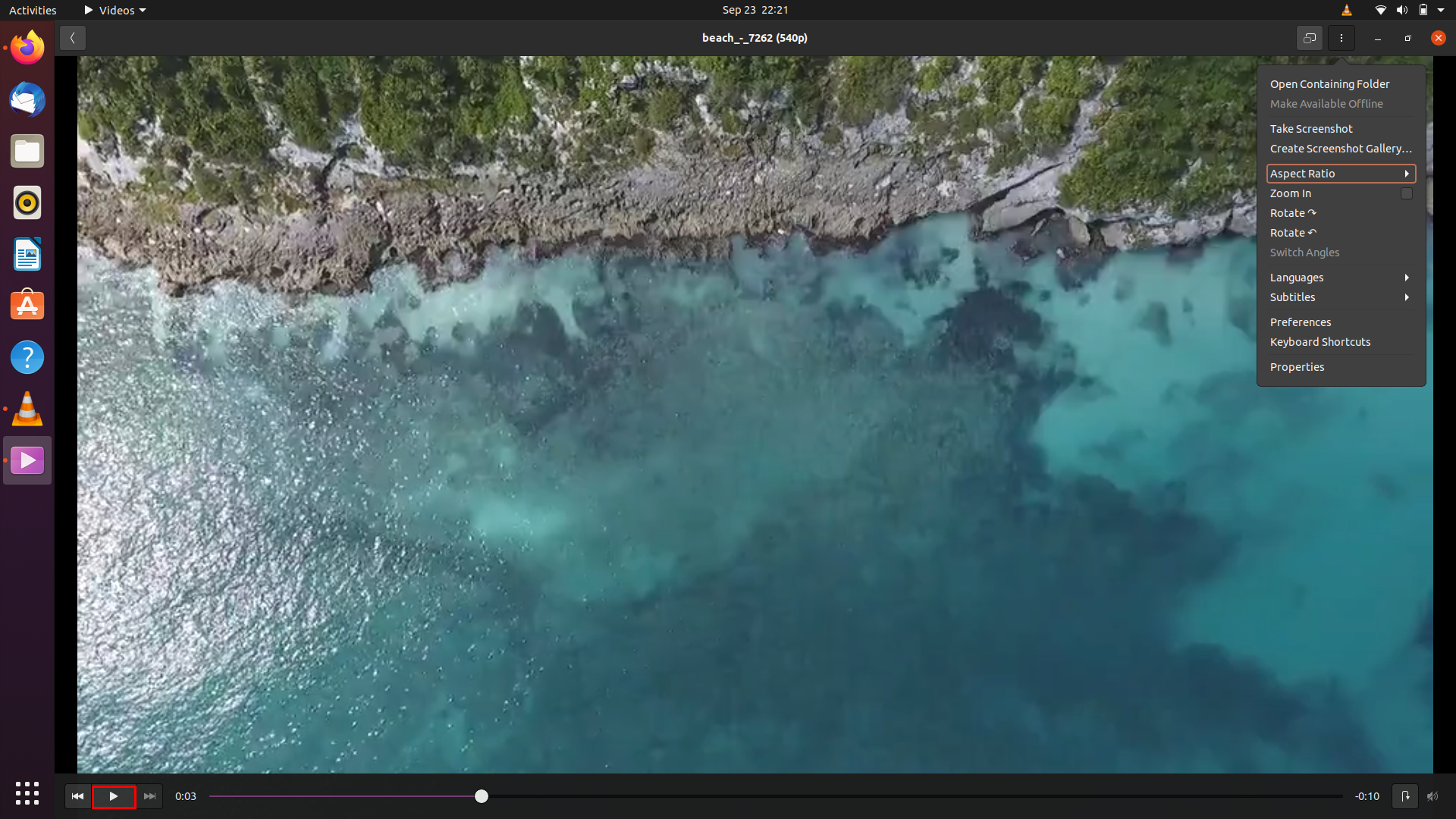}
        \caption{Example of VLC. \textbf{Original:} ‘Play’;
            \textcolor{darkgreen}{\textbf{Appearance:} “White triangle icon within a black circular button.”};
            \textcolor{darkred}{\textbf{Spatial:} “Located at the bottom left corner of the screen.”};
            \textcolor{darkblue}{\textbf{Intention:} “Click to play the video.”}}
    \end{subfigure}

    \vspace{2em} 

    \begin{subfigure}{0.49\textwidth}
        \centering
        \includegraphics[width=\textwidth]{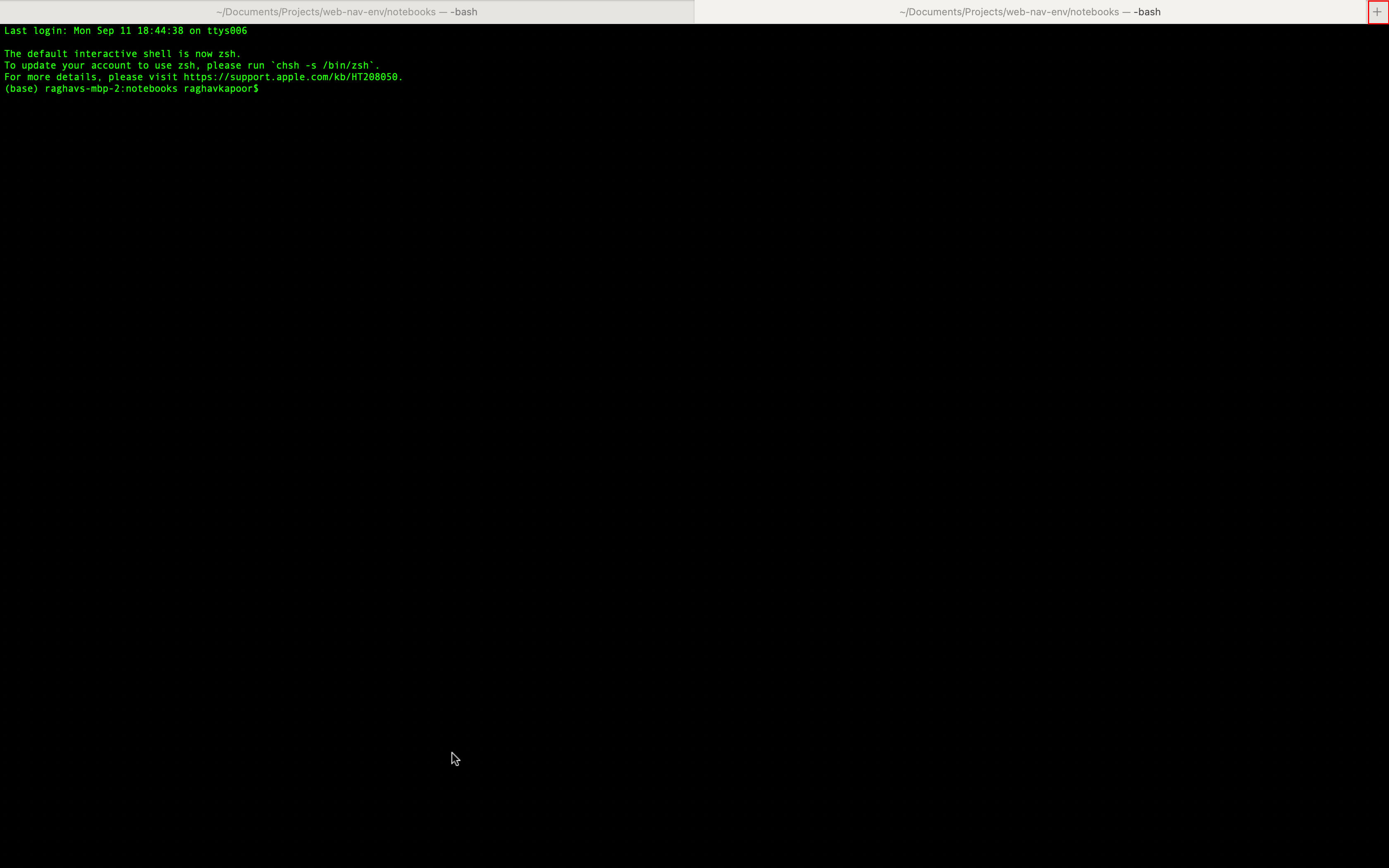}
        \caption{Example of Terminal. \textbf{Original:} ‘create\_new\_tab’;
            \textcolor{darkgreen}{\textbf{Appearance:} “A small '+' icon in a gray tab bar.”};
            \textcolor{darkred}{\textbf{Spatial:} “Located at the far right of the tab bar.”};
            \textcolor{darkblue}{\textbf{Intention:} “Open a new terminal tab.”}}
    \end{subfigure}
    \hfill
    \begin{subfigure}{0.49\textwidth}
        \centering
        \includegraphics[width=\textwidth]{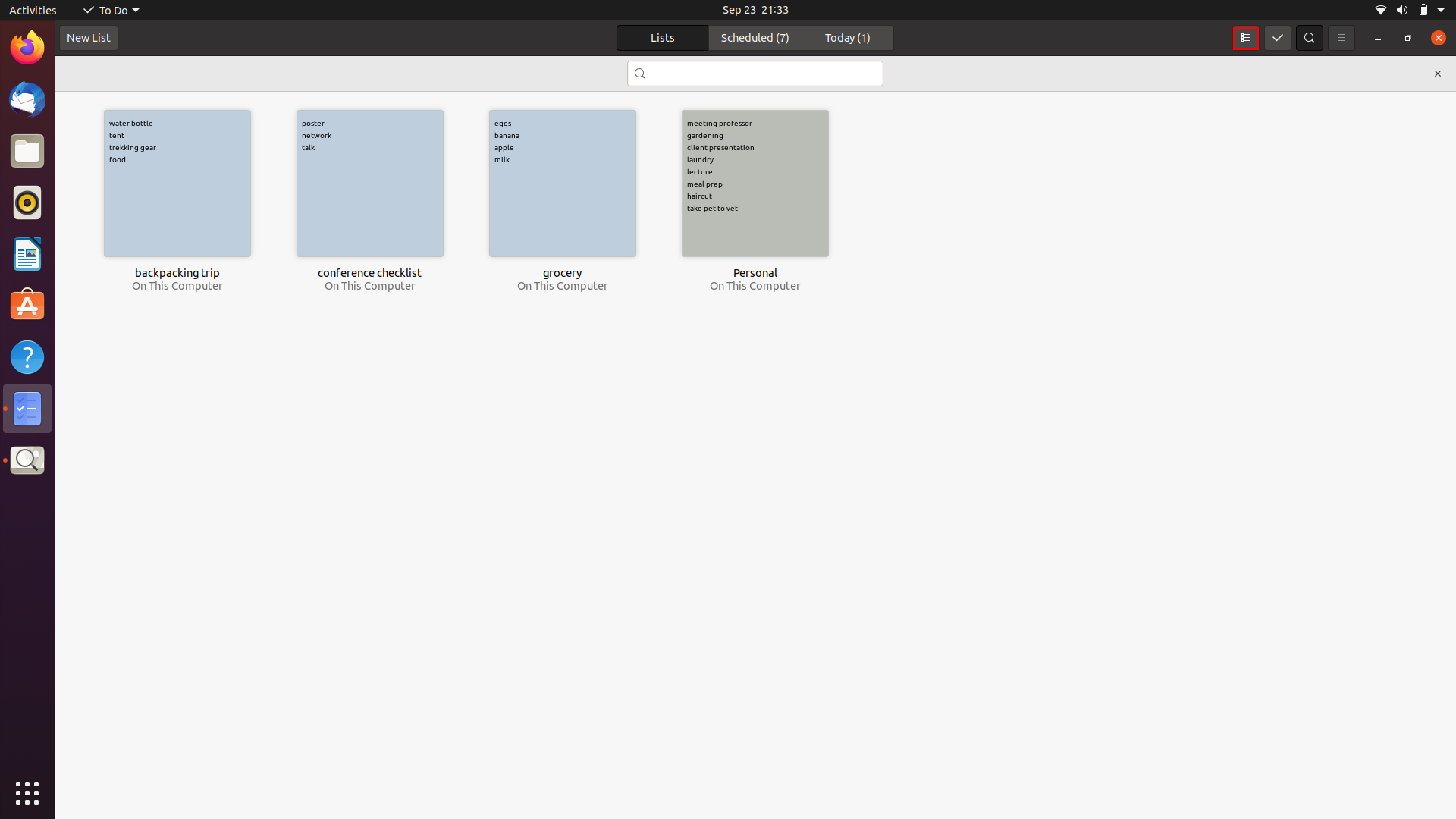}
        \caption{Example of Todo. \textbf{Original:} ‘view as list’;
            \textcolor{darkgreen}{\textbf{Appearance:} “A gray, vertical button with a box and lines icon.”};
            \textcolor{darkred}{\textbf{Spatial:} “Positioned at the top right beside the search bar.”};
            \textcolor{darkblue}{\textbf{Intention:} “Switch to list view.”}}
    \end{subfigure}
    \vspace{2em} 

    \caption{Illustration of how we augment the original OmniAct-Desktop annotations with diverse queries based on Appearance, Spatial Relationships, and Intention.}
    \label{supp:fig:gpt4o}
\end{figure*}

In Fig.\ref{supp:fig:gpt4o}, we demonstrate more examples about how we leverage GPT4o to augment the original OmniAct-Desktop annotations with diverse queries based on Appearance, Spatial Relationships, and Intention.

\noindent\textbf{(iii) Mobile--Functionality:} mobile data is readily available in Android like~\cite{ricosca, aitw}, which provide icon captioning. Notably, we consider~\cite{amex} as its valuable functionality descriptions that go beyond simple atomic element names.

\noindent\textbf{Balance Data by Sampling:}
As shown in Tab.\ref{table:dataset}, data scale varies significantly across types (\eg, only 100 desktop samples). To ensure fair exposure over each type, we develop a sampling strategy during training, giving each batch a comparable probability of including different data types.

\begin{table}[t!]
\footnotesize
    \tablestyle{1.2pt}{0.98}
    \resizebox{0.48\textwidth}{!}{   
    \begin{tabular}{ccccccc}
        \toprule
        \textbf{Usage} & \textbf{Device} & \textbf{Source} & \textbf{\#Sample} & \textbf{\#Ele.} & \textbf{\#Cls. (len.)} & \textbf{Highlights} \\
        \midrule
        \multirow{3}{*}{\textbf{Grounding}} & Web & Self-collected & 22K & 576K & N/A & Visual-based \\
        & Mobile & AMEX~\cite{amex} & 97K & 926K & N/A &  Functionality \\
        & Desktop & OmniAct~\cite{omniact} & 100 & 8K & N/A &  Diverse query \\
        \midrule
        \multirow{2}{*}{\textbf{Navigation}} & Web & GUIAct~\cite{guicourse} & 72K & 569K &  9 (7.9) & One / Multi-step\\
        & Mobile & GUIAct~\cite{guicourse} & 65K & 585K & 5 (9.0) & Multi-step\\
        \midrule
        \textbf{Total} & Diverse &  & 256K & 2.7M & \\  
        \bottomrule        
    \end{tabular}
    }
    \caption{
    Overview of our instruction-tuning data. 
    \textbf{\#Sample} indicates the number of the task instance (screenshot in grounding, task in navigation);
    \textbf{\#Ele.} indicates the number of the element (\ie, bbox in grounding);
    \textbf{\#Cls.} represents the number of action classes, and \textbf{len.} indicates the average trajectory length per task.
    }
    \label{table:dataset}
\end{table}

\section{Experiments}

\subsection{Benchmark Datasets}
We evaluate our model using the following benchmarks:

\noindent\textbf{Grounding.} We use Screenspot~\cite{seeclick}, a zero-shot grounding evaluation benchmark with diverse data across three device, to assess text and widget recognition separately.

\noindent\textbf{Navigation.} We evaluate navigation performance across four different datasets from different devices:
{{(\textit{i})} Web} on Mind2Web~\cite{mind2web}, which includes an action space with three types of actions.
{(\textit{ii}) Mobile} on AITW~\cite{aitw}, which action spaces includes 11 actions.
{(\textit{iii}) Online} on MiniWob~\cite{miniwob++}, an online environment with two types of actions, introduced to complement the offline benchmarks and test performance in an interactive setting.
Details of the training settings are provided in the supplementary material.


\subsection{Main Results}
We organize our experiments on each downstream tasks to address the following questions: 
\textbf{Q1:} How does our model perform on each task? What improvements are achieved beyond existing VLM baseline?
\textbf{Q2:} What are the effects and improvements of each component?
\textbf{Q3:} What insights can be gained from each benchmark based on its unique properties?


\subsubsection{Grounding Tasks}

\begin{table}[!h]\centering
\tablestyle{2pt}{0.98}
\resizebox{\columnwidth}{!}
{
\begin{tabular}{llcrrrrrrrr}\toprule
\multirow{2}{*}{Method} &\multirow{2}{*}{Size} &\multirow{2}{*}{\#Train} &\multicolumn{2}{c}{{Mobile}} &\multicolumn{2}{c}{{Desktop}} &\multicolumn{2}{c}{{Web}} &\multirow{2}{*}{\textbf{Avg.}} \\
& & &Text &Icon &Text &Icon &Text &Icon & \\\midrule
\base~\cite{qwen2vl} & 2B & -- & 24.2 & 10.0 & 1.4 & 9.3 & 8.7 & 2.4 & 9.3\\
Fuyu~\cite{fuyu-8b} & 8B &--&41.0 &1.3 &33.0 &3.6 &33.9 &4.4 &19.5 \\
CogAgent~\cite{cogagent} &18B & 400K &67.0 &24.0 &74.2 &20.0 &{70.4} &28.6 &47.4 \\
SeeClick~\cite{seeclick} & 9.6B & 364K & {78.0} &52.0 &72.2 &30.0 &55.7 &32.5 &53.4 \\
\rowcolor[gray]{0.9} OmniParser~\cite{omniparser} & \small * & -- & 93.9 & 57.0 & 91.3 & 63.6 & 81.3 & 51.0 & 73.0\\
UGround~\cite{uground} & 7B & 1.3M & 82.8 & 60.3 & 82.5 & 63.6 & 80.4 & 70.4 & 73.3\\
\midrule
\our-G & {2B} & 119K & 91.6 & 69.0 & 81.8 & 59.0 & 83.0 & 65.5 & {74.9} \\
\our & {2B} & 256K & 92.3 & 75.5 & 76.3 & 61.1 & 81.7 & 63.6 & \textbf{75.1} \\
\bottomrule
\end{tabular}
}
\caption{
{Zero-shot grounding on \screenspot}.
* means Omniparser use GPT-4V. ``Size'' refers to model size.
\our-G: trained solely on grounding data, excluding navigation data.
\our, delivers strong grounding results with a lightweight model size and minimal training data. 
}
\label{tab:screenspot}
\end{table}
In Tab.~\ref{tab:screenspot}, we present zero-shot grounding results on the \screenspot~\cite{seeclick}.
This provides a straightforward signal of the shortcomings in each setup. 
We includes one additional variant -- \our-G, which only used grounding data for training.
Our finding includes:
\textbf{(i)} Overall all methods, the text track scores are higher than the icon track, even for desktop text, which was less seen during training. This suggests that text grounding ability mainly learned from web and mobile is transferable across platforms.
\textbf{(ii)} 
Mixing navigation data~\cite{guicourse} does not impair grounding performance when an effective sampling strategy is used.
\textbf{(iii)} The icon track is more challenging due to its visual grounding. Mobile scores are significantly higher than desktop and web, this emphasize the \textbf{missing of visual UI grounding data beyond mobile devices}.
\textbf{(iv)} Notably, \textbf{\our, as the most lightweight method with minimal training data}, achieves state-of-the-art grounding performance.


\subsubsection{Navigation Tasks}
\noindent\textit{Mobile: AITW.}
In Tab.\ref{tab:aitw}, we have the following findings:
\textbf{(i)} Without interleaved streaming (i.e., without visual history), \our$\dagger$ provides only a 1.1\% acc. improvement over the VLM baseline. However, with visual history, \our~achieves an additional 1.7\% acc. gain, likely because \textbf{visual context is crucial} for adapting to frequent software changes within a large action space (11), particularly on mobile platforms.
\textbf{(ii)} \textbf{\our's zero-shot navigation learned from GUIAct~\cite{guicourse} demonstrates transferability}, suggesting further improvements can be made to the navigation component.
\textbf{(iii)} \our, beats OmniParser~\cite{omniparser} and PaLM2-CoT~\cite{palm2-cot}, which either leverage closed-source APIs or HTML information, highlighting its \textbf{potential of a single model as a standalone visual agent.}

\begin{table}[!h]\centering
\tablestyle{1.8pt}{0.98}
\resizebox{\columnwidth}{!}
{
\begin{tabular}{lcrrrrrrrr}\toprule
Method &FT? &General &Install &G.Apps & Single & WebShop. &\textbf{Overall} \\\midrule
\rowcolor[gray]{0.9}
ChatGPT-CoT~\cite{chatgpt-cot} & -- &5.9 &4.4 &10.5 &9.4 &8.4 &7.7 \\
\rowcolor[gray]{0.9}
PaLM2-CoT~\cite{palm2-cot} &-- &-- &-- &-- &-- &-- &39.6 \\
\rowcolor[gray]{0.9}OmniParser~\cite{omniparser} & * & 48.3 & 57.8 & 51.6 & 77.4 & 52.9 & 57.7\\
SeeClick~\cite{seeclick}  & \cmark &54.0 &66.4 &54.9 &63.5 &57.6 &59.3 \\
\baseshort~\cite{qwen2vl}  & \cmark & 61.4 & 71.8 & 62.6 & 73.7 & 66.7 & 67.2 \\
\midrule
\our$\dagger$ & \cmark & 63.5 & 72.3 & 66.0 & 72.3 & 65.8 & {68.3}\\
\our  & \cmark & 63.9 & 72.5 & 69.7 & 77.5 & 66.6  & \textbf{70.0}\\
\our-ZS  & \xmark & 32.1 & 47.7 & 42.0 & 20.1 & 37.4 & 35.9 \\
\bottomrule
\end{tabular}
}
\caption{{Performance of Mobile Navigation~\cite{aitw}},
where \colorbox{gray!20}{\makebox(15,4){gray}} color indicates these methods either using HTML inputs or rely on close-source GPT-4V (marked with *).
\our$\dagger$ denotes our variant without interleaved visual-action streaming, utilizing only action history.
}
\label{tab:aitw}
\end{table}


\noindent\textit{Website: Mind2Web.}
In Tab.~\ref{tab:mind2web} for web navigation, we found that:
\textbf{(i)} Instruction-tuning has a significant effect, brings 4.6\% Avg. Step SR. boost over \base. Notably, \textbf{\our-2B's zero-shot yield comparable with SeeClick-9.6B which has pretrained and fine-tuning}, and achieves relatively high Op. F1 (80\%+).
\textbf{(ii)} Visual context in this task is less significant than in AITW, possibly because Mind2Web focuses primarily on a single, visually similar website and includes only three actions. 
\textbf{(iii)} The {cross-website and cross-domain settings are harder than cross-tasks}. This suggests the \textbf{bottleneck is lie in UI visual perception rather than textual task understanding} (websites/domains are unseen in training).
One future effort for improvement is to \textbf{develop training data with good (visually) domain diversity.}

\noindent\textit{Online: MiniWob.}
In Tab.~\ref{tab:miniwob}, this benchmark demonstrates model behavior in an online environment.
Our key finding is that despite the simplicity of the Miniwob UI, the gap between \our's zero-shot performance (27.1\%) and fine-tuned Qwen-VL (48.4\%) is substantial. In contrast, \our's zero-shot surpasses in Mind2Web, likely because out-of-distribution errors are not adequately addressed in the instruction-tuning phase. This suggests that offline instruction-tuning alone is insufficient; we need to \textbf{develop a learning strategy tailored for an online environment}, which can handle novel error cases.

\begin{table*}[!t]
\centering
\begin{minipage}{0.73\textwidth}
\centering
\tablestyle{3pt}{0.85}
\resizebox{\textwidth}{!}{
\begin{tabular}{llccccccccc}\toprule
\multirow{2}{*}{Method} &\multirow{2}{*}{FT?} &\multicolumn{3}{c}{\textbf{Cross-Task}} &\multicolumn{3}{c}{\textbf{Cross-Website}} &\multicolumn{3}{c}{\textbf{Cross-Domain}} \\
\cmidrule{3-11}
& & Ele.Acc &Op.F1 &Step.SR &Ele.Acc &Op.F1 &Step.SR &Ele.Acc &Op.F1 &Step.SR \\\midrule
\rowcolor[gray]{0.9}
MindAct~\cite{mind2web} &-- &55.1 &75.7 &52.0 &42.0 &65.2 &38.9 &42.1 &66.5 &39.6  \\
\rowcolor[gray]{0.9}
GPT-4~\cite{gpt4} &-- &41.6 &60.6 &36.2 &35.8 &51.1 &30.1 &37.1 &46.5 &26.4 \\
\rowcolor[gray]{0.9}
OmniParser~\cite{omniparser} & * & 42.4 & 87.6 & 39.4 & 41.0 & 84.8 & 36.5 & 45.5 & 85.7 & 42.0\\
CogAgent~\cite{cogagent} & \cmark & 22.4 & 53.0 & 17.6 & 18.4 & 42.4 & 13.4 &  20.6 & 42.0 & 15.5\\
Qwen-VL~\cite{Qwen_technicalReport}& \cmark &15.9 &86.7 &13.3 &13.2 &83.5 &9.2 &14.1 &84.3 &12.0 \\
SeeClick~\cite{seeclick} & \cmark &28.3 &87.0 &25.5 &21.4 &80.6 &16.4 &23.2 & {84.8} &20.8  \\
\base~\cite{qwen2vl} & \cmark & 37.7 &86.4 & 33.2 & 36.0 & 79.2 & 27.6 & 36.3 & 81.8 & 30.7 \\
\midrule
\our$\dagger$  & \cmark & 39.7 & 88.0 & 36.9 & 41.0 & \textbf{83.6} & 34.2 & 38.9 & 85.3 & 34.1 \\
\our  & \cmark & \textbf{39.9} & \textbf{88.6} & \textbf{37.2} & \textbf{41.6} & 83.5 & \textbf{35.1} & \textbf{39.4} & \textbf{86.8} & \textbf{35.2}\\
\our-ZS  & \xmark & 21.4 & 85.2 & 18.6 &21.9 & 81.9 & 16.8 & 24.4 & 83.9 & 21.4 \\
\bottomrule
\end{tabular}
}
\caption{{Web Navigation on \mindweb.}
The \colorbox{gray!20}{\makebox(15,4){gray}} correspond to methods that required HTML text inputsor rely on close-source GPT-4V (marked with *).
\our$\dagger$ denotes our variant utilizing only action history.
{\our's zero-shot performance yield comparable score with SeeClick with pretrained and fine-tuning}.
}

\label{tab:mind2web}
\end{minipage}%
\hfill
\begin{minipage}{0.25\textwidth}
\centering
\tablestyle{5pt}{1.1}
\resizebox{\textwidth}{!}{
\begin{tabular}{lcc}\toprule
Method & FT? & Score \\ \midrule
CC-Net(SL)~\cite{humphreys2022data_driven}  & \cmark & 23.4 \\
Pix2Act~\cite{pixel2act}  & \cmark & 55.2 \\
Qwen-VL~\cite{Qwen_technicalReport}  & \cmark & 48.4 \\
SeeClick~\cite{seeclick}  & \cmark & 67.0 \\
\base~\cite{qwen2vl} & \cmark & 66.8\\
\midrule
\our$\dagger$  & \cmark & {70.4}\\
\our  & \cmark & \textbf{71.5}\\
\our-ZS  & \xmark & 27.1\\
\bottomrule
\end{tabular}
}
\caption{Results on online navigation MiniWob on 35-tasks split following SeeClick~\cite{seeclick}.
\our$\dagger$ denotes our variant utilizing only action history.
}
\label{tab:miniwob}
\end{minipage}
\end{table*}

\begin{figure}[!h]
\centering
\begin{minipage}[t]{1.0\linewidth}
\centering
\tablestyle{1.5pt}{0.98}
\resizebox{\linewidth}{!}{
\begin{tabular}{lrcccc}
\toprule
\textbf{Method} & \textbf{Strategy} & \textbf{\#Vis.Ctx.} & \textbf{Train.Speedup} & \textbf{Test-time?} & \textbf{Screenspot}  \\ 
\midrule
Baseline & N/A & 1344.0 & 1$\times$ & N/A & 70.8 \\
\midrule
\multirow{2}{*}{Token Merge} & \multirow{2}{*}{UI-Graph} & \multirow{2}{*}{852.8} & \multirow{2}{*}{1.6$\times$} &  & 42.3\\
 & &  & & \cmark & 34.7\\
\midrule
\multirow{4}{*}{Token Selection}  
 & \multirow{2}{*}{Random} & \multirow{2}{*}{941.5} & \multirow{2}{*}{1.5$\times$} &  & 65.3\\
 & &  & & \cmark & 56.2\\
 \cmidrule{2-6}
 & \multirow{2}{*}{UI-Graph} & \multirow{2}{*}{947.4} & \multirow{2}{*}{1.5$\times$} & & \textbf{70.4}\\
 & &  & & \cmark & 64.9\\
\bottomrule
\end{tabular}
}
\subcaption{Comparison between different visual tokens compression methods.
‘\#Vis.ctx' represents the avg. length of visual tokens across all layers. 
‘Train.Speedup’ denotes the training efficiency improvement over the baseline. 
‘Inference’ denotes whether this method is applicable at test time.
}
\label{tab:ablation:select}
\end{minipage}%
 \vspace{1em}
\hfill
\begin{minipage}[t]{0.45\linewidth}
\centering
\tablestyle{1.8pt}{1.1}
\resizebox{\linewidth}{!}
{
\begin{tabular}{lcc}
\toprule
\textbf{Insertion}  & \textbf{\#layers} & \textbf{Screenspot} \\ 
\midrule
All &28 & 65.7\\
Early & 14 & 68.2\\
Late & 14 & 67.6 \\
Cross & 14& 70.5\\
\bottomrule
\end{tabular}
}
\subcaption{Different insertion layers.}
\label{tab:ablation:insert}
\end{minipage}
\hfill
\begin{minipage}[t]{0.45\linewidth}
\centering
\tablestyle{1.5pt}{0.98}

\resizebox{\linewidth}{!}
{
\begin{tabular}{lrc}
\toprule
\textbf{Ratio}  & \textbf{\#Visual Ctx.} & \textbf{Screenspot} \\ 
\midrule
0 & 1344.0 & 70.8 \\
0.25 & 1185.2 & 70.6\\
0.5 & 947.4 & 70.4\\
0.75 & 848.6 & 68.3\\
1.0 & 762.1 & 64.5\\
\bottomrule
\end{tabular}
}
\subcaption{Different selection ratio.}
\label{tab:ablation:ratio}
\vfill
\end{minipage}
 \vspace{1em}    
\caption{Ablation studies of several design chocies regarding our UI-Guided Token Selection.
}
\label{tab:ablation:ui}
\end{figure}

\subsection{Ablation Studies}
\noindent\textbf{Impact by UI-Guided Token Selection.}
In Tab.\ref{tab:ablation:select}, we examine various visual token optimization methods through the following variants:
{(i) Baseline:} no visual token optimization strategy applied;
{(ii) Token Merge:} a mainstream method introduced in Sec.\ref{sec:uiguided}, conditioned on our UI-Graph;
{(iii) Token Select.-Random:} a variant that randomly selects a subset of visual tokens, serving as a direct baseline;
{(iv) Token Select.-UI-Graph:} our proposed method leveraging UI-Graph for token selection.

As shown, Token Merge performs worse than random selection, highlighting the importance of preserving positional relationships between tokens. Token Selection - UI-Graph offers a good balance with a 1.5× speedup and competitive accuracy. While applying it at test-time slightly reduce accuracy due to resolution loss, it remains more reliable than random selection, demonstrating the effectiveness of UI connected graph as a guiding criterion.

\noindent \textbf{Selection of Layers Insertion.}
In Tab.\ref{tab:ablation:insert}, we present an ablation study on different insertion strategies, including insertion across all layers, early or late X layers, and {cross}-layer insertion, where layers alternate between inserted and non-inserted. With the same number of inserted layers, cross-layer insertion significantly outperforms both early and late insertion. 

\noindent \textbf{Different Selection Ratio.} The results is present in Tab.\ref{tab:ablation:ratio}, illustrate the selection ratio as a trade-off between speedup and performance, with 0.5 as an effective choice.

\begin{figure}[h]
    \centering
    \begin{minipage}{0.48\linewidth}
        \centering
        \includegraphics[width=\textwidth]{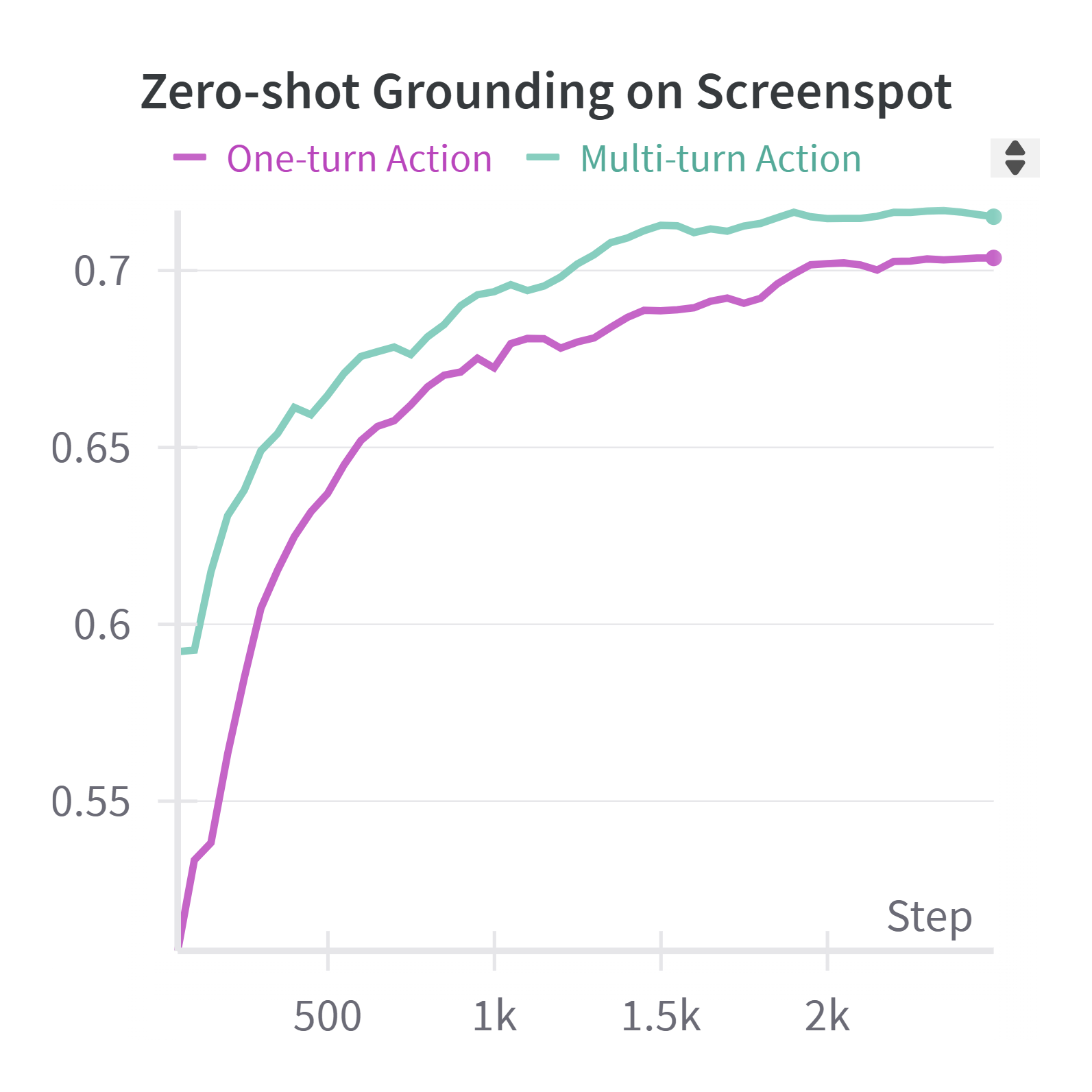}
        \caption{Impact by Interleaved action-query streaming on Grounding task: trained with 119K grounding data, Eval with Screenspot.}
        \label{fig:stream:grd}
    \end{minipage}
    \hfill
    \begin{minipage}{0.48\linewidth}
        \centering
        \includegraphics[width=\textwidth]{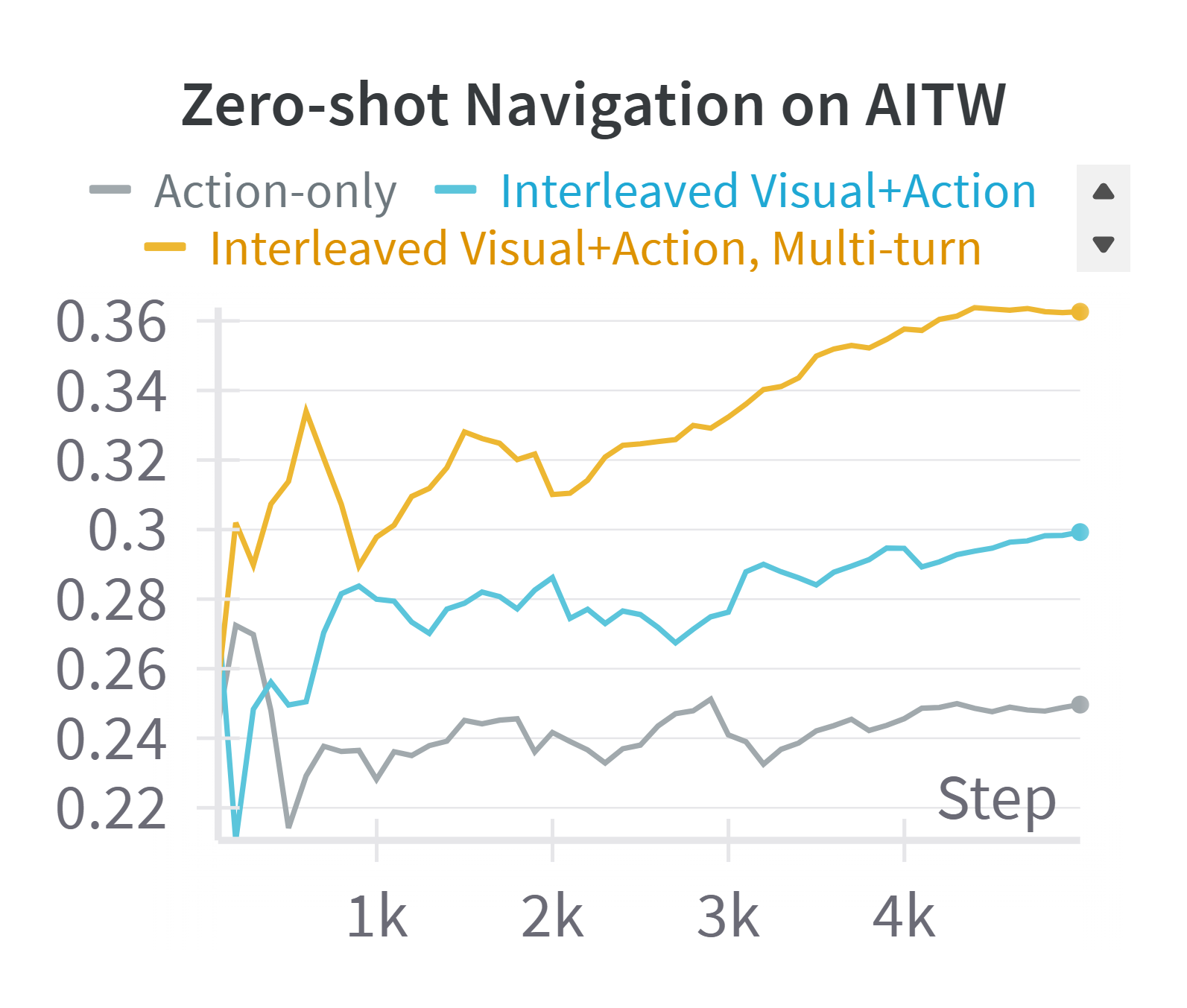}
        \caption{Impact by Interleaved action-visual streaming on Navigation task: trained with GUIAct, Eval with AITW.}
        \label{fig:stream:nav}
    \end{minipage}
\end{figure}

\noindent \textbf{Impact of Interleaved Streaming.}
We evaluate performance over iterations using interleaved streaming for grounding and navigation tasks to study its effects.
\textbf{(i) Action-Query:} In Fig.~\ref{fig:stream:grd}, we compare grounding training with and without multi-turn streaming. Multi-turn streaming leads to faster progress, especially in the initial warmup phase, and maintains a performance gap, demonstrating improved data utilization.
\textbf{(ii) Action-Visual:} As shown in previous tables with the \our$\dagger$ variant, we validated the impact by visual context. Fig.~\ref{fig:stream:nav} illustrates model progress across iteration steps, where the trend shows that visual+action+multi-turn outperforms visual-action and action-only setups. 
This validates our interleaved streaming as an effective and efficient strategy.

\noindent\textbf{D. Impact by Instruction-Tuning Data.}
One our contributions is the analysis of training data for the grounding task in Sec.\ref{sec:data}.
In Tab.\ref{abl:grounding}, we present a detailed ablation study to investigate the individual impact of each change on specific devices and settings.

\begin{table}[!h]
\centering
\tablestyle{1.5pt}{0.98}
\resizebox{\columnwidth}{!}{
\begin{tabular}{lccrrrrrrrr}
\toprule
\multirow{2}{*}{\textbf{Training Data}} &\multirow{2}{*}{\textbf{\#Sample}} &\multirow{2}{*}{\textbf{\#Ele.}} &\multicolumn{2}{c}{{Mobile}} &\multicolumn{2}{c}{{Desktop}} &\multicolumn{2}{c}{{Web}} &\multirow{2}{*}{\textbf{Avg.}} \\
& & &Text &Icon &Text &Icon &Text &Icon & \\\midrule
AMEX & 97K & 1.06M & 90.1& 66.8 & 78.8 & 50.7 &78.3& 55.3 &  70.0\\
Web (Seeclick~\cite{seeclick}) & 270K & 3.0M & 83.9 & 61.1 & 70.6 & 47.9 & 73.0 & 56.3 & 65.5 \\
Web (text+vis) & 22K & 926K & 85.7 & 60.7 & 75.8 & 50.0 & 74.4 & 52.9 & 66.6\\
Web (vis) & 22K & 576K & 83.6 & 60.3 & 75.3 & 60.3 & 72.4 & 62.1 & 69.0 \\
OmniAct & 100 & 2K & 85.7 & 59.8 & 78.4 & 52.1 & 79.6 & 52.9 & 68.7\\
OmniAct (diverse) & 100 & 8K & 88.3 & 66.8 & 76.3 & 57.9 & 79.1 & 57.3 & 70.9\\
\midrule
Joint-Training & 119K & 1.6M & 90.8 & 66.4& 75.3 & 51.4 & 80.0 & 63.1 & 71.2 \\
\textbf{Balanced Sampling} & 119K & 1.6M & 91.6 & 69.0 & 81.8 & 59.0 & 83.0 & 65.5 & \textbf{74.9} \\
\bottomrule
\end{tabular}
}
\caption{Effect by individual training data on Screenspot.}
\label{abl:grounding}
\end{table}
We found that 
\textbf{(i) Data quality matters:} OmniAct, with only 2K elements, achieves comparable scores to web data, and when augmented with GPT4o for diverse queries, it enhances model generalization and optimizes data usage efficiency.
\textbf{(ii)} Our collected 22K web data outperforms SeeClick's 270K screenshots; additionally, filtering web data by visual criteria significantly reduces element size without impacting performance, suggesting that static text is less informative—a property inherent to VLM.
\textbf{(iii) Balanced sampling is essential,} yielding a 3.7\% acc. gain and maintaining performance across individual setups.

\subsection{Qualitative Examples}
\begin{figure*}[htbp]
    \centering
    \begin{subfigure}{0.58\textwidth}
        \centering
        \includegraphics[width=\textwidth]{figures/screenspot/True_desktop_icon_screenspot_216_open_wechat.png}
        \caption{
        \textbf{\textcolor{darkgreen}{\cmark} Instruction: “Open wechat”.}
With instruction-tuning, \our~can recognize the appearance of the WeChat icon.
}
    \end{subfigure}
    \hfill
    \begin{subfigure}{0.4\textwidth}
        \centering
        \includegraphics[width=\textwidth]{figures/screenspot/True_desktop_icon_screenspot_148_rotate_left.png}
        \caption{
        \textbf{\textcolor{darkgreen}{\cmark} Instruction: “Rotate left”.}
\our~distinguishes the correct operator among multiple abstract symbols.
}
    \end{subfigure}

    \vspace{2em} 
    \begin{subfigure}{0.49\textwidth}
        \centering
        \includegraphics[width=\textwidth]{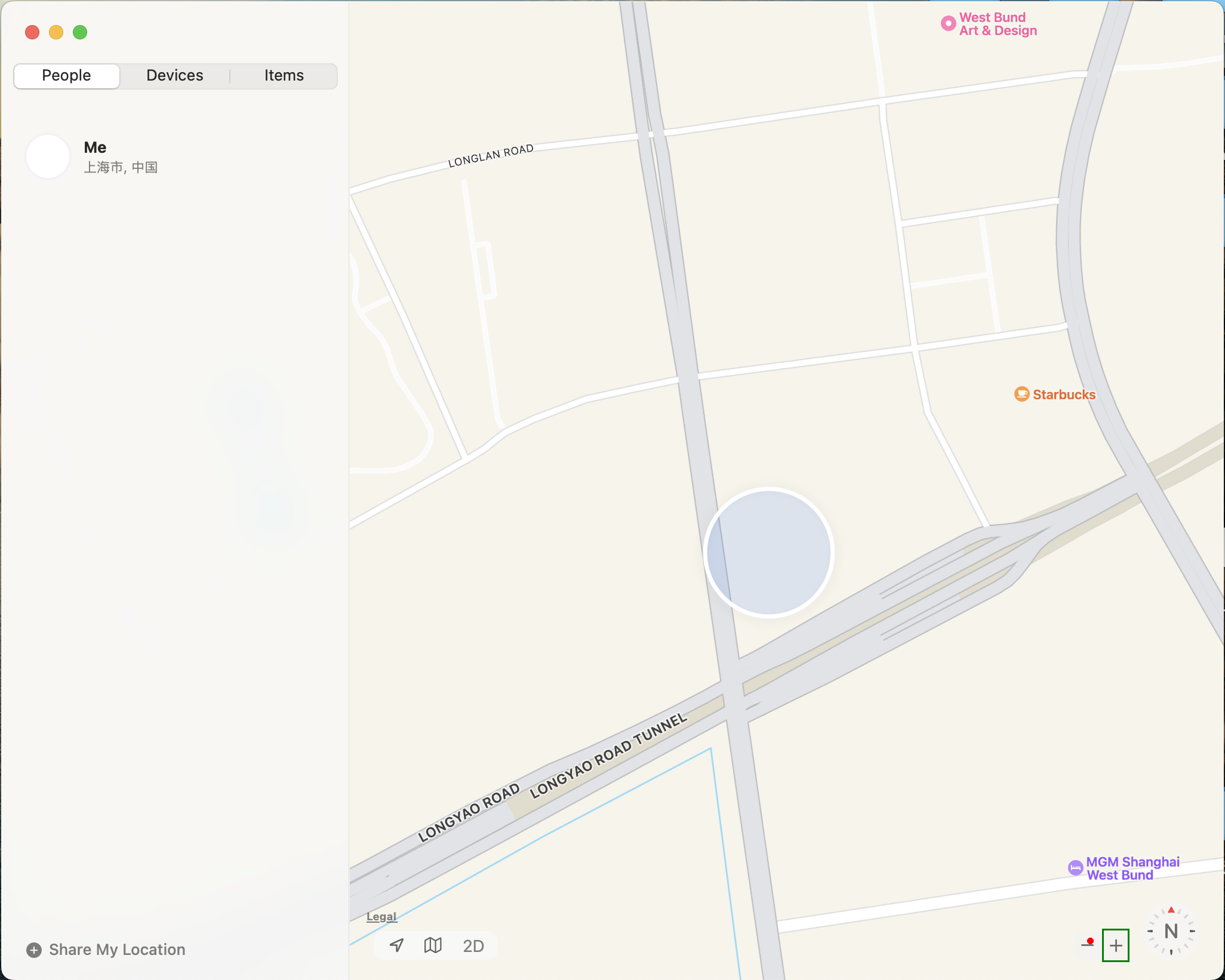}
        \caption{
        \textbf{\textcolor{darkred}{\xmark} Instruction: “Zoom in”.}
        The model visually confuses the difference between zoom in and zoom out.}
    \end{subfigure}
    \hfill
    \begin{subfigure}{0.49\textwidth}
        \centering
        \includegraphics[width=\textwidth]{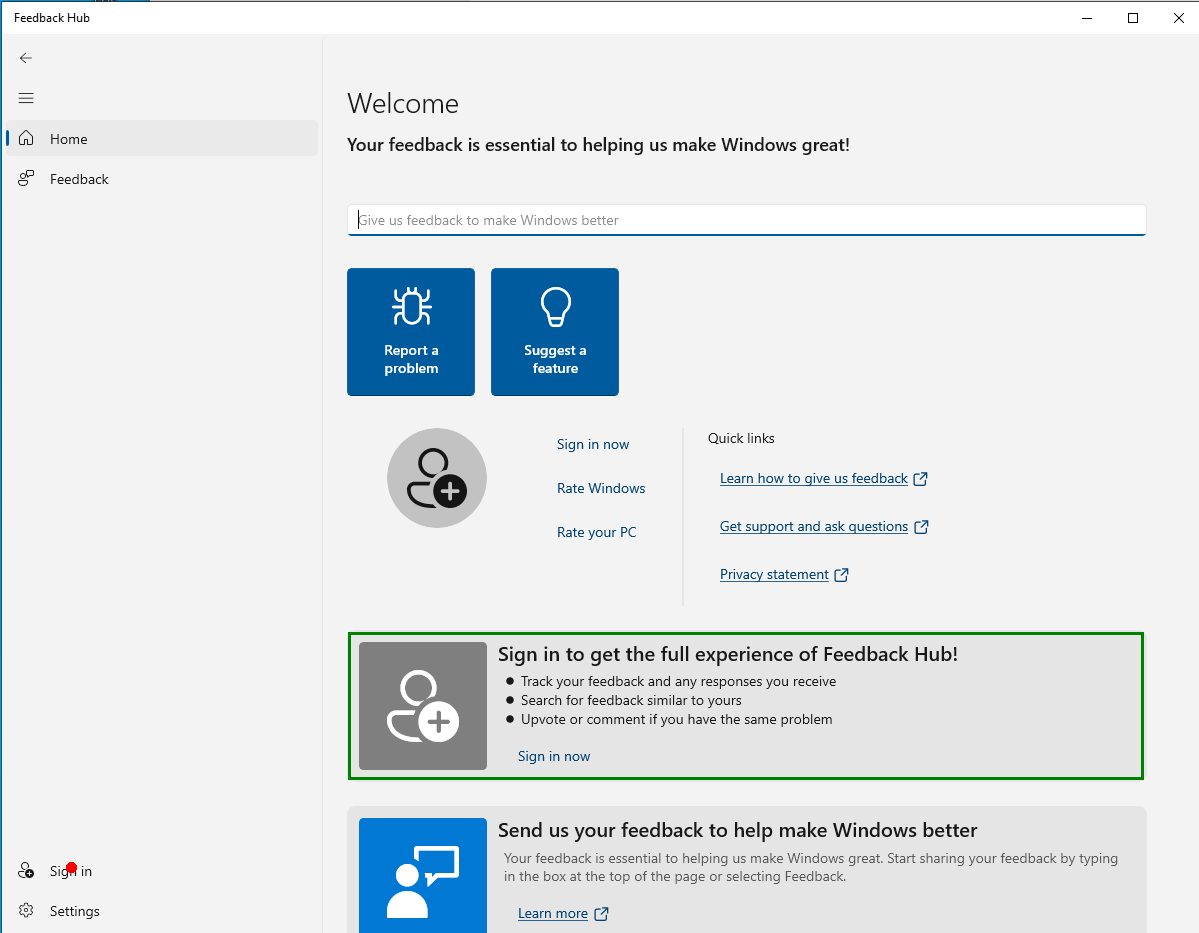}
        \caption{
        \textbf{\textcolor{darkred}{\xmark} Instruction: “Sign in”.}
        There are two possible sign-in elements, but the query lacks sufficient information to determine the correct one.}
    \end{subfigure}

    \vspace{2em} 
    \begin{subfigure}{0.49\textwidth}
        \centering
        \includegraphics[width=\textwidth]{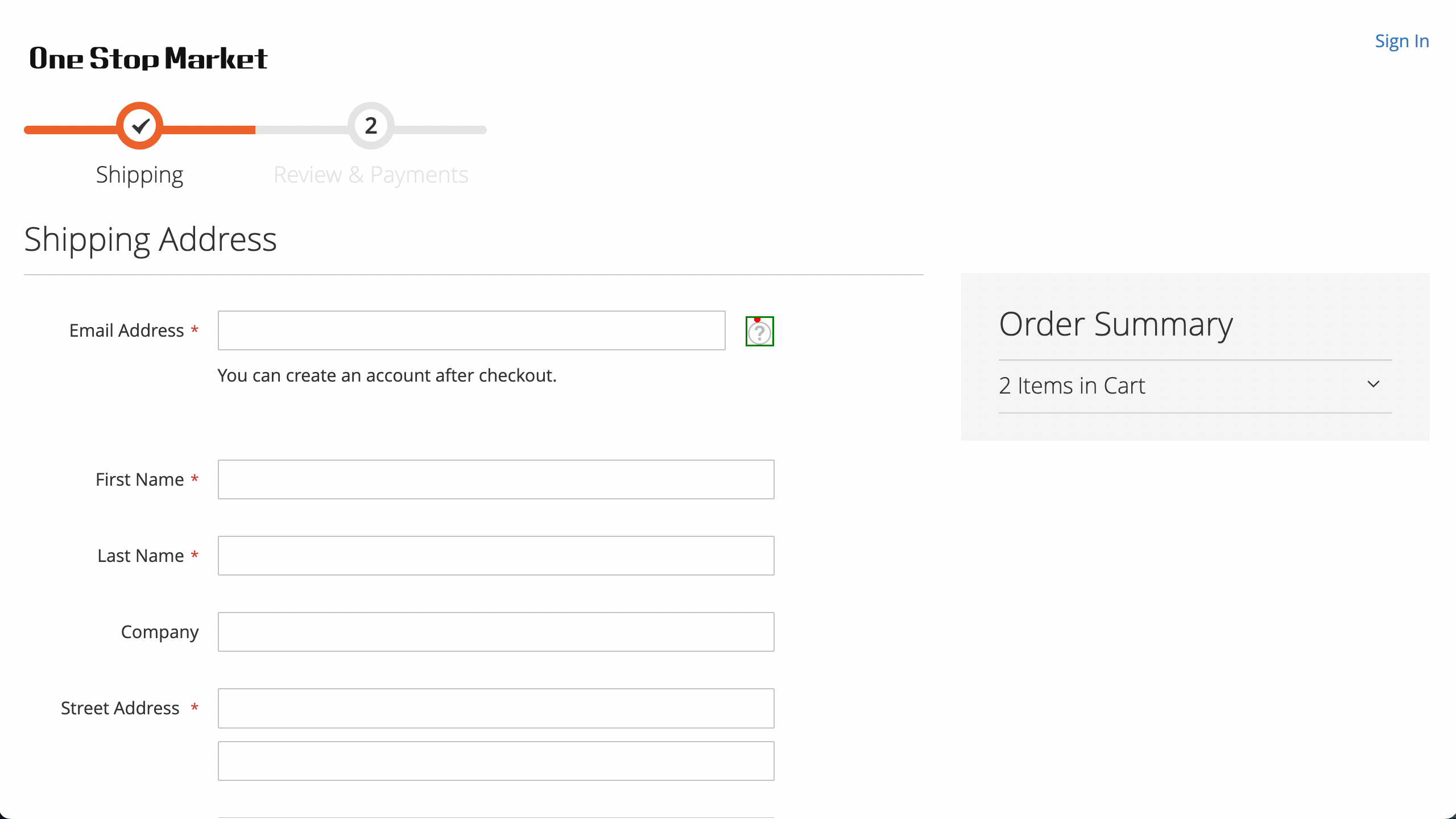}
        \caption{
        \textbf{\textcolor{darkgreen}{\cmark} Instruction: “view help for email account”.}
\our~is able to associate “view help” with question mark clickable element.
}
    \end{subfigure}
    \begin{subfigure}{0.49\textwidth}
        \centering
        \includegraphics[width=\textwidth]{figures/screenspot/False_web_icon_screenspot_1_view_my_account.png}
        \caption{
        \textbf{\textcolor{darkred}{\xmark} Instruction: “view my account”.}
‘View my account’ could be interpreted as ‘Click Your Profile’ or ‘User Profile’ (top right), leading to confusion.
}
    \end{subfigure}
    \hfill
    
    \vspace{2em}
    \caption{Case studies on Screenspot's Desktop (a-d) and Web (e-f) grounding,
    } 
    \label{supp:fig:screenshot:desktop}    
\end{figure*}

\begin{figure*}[htbp]
    \centering
    \begin{subfigure}{0.24\textwidth}
        \centering
        \includegraphics[width=\textwidth]{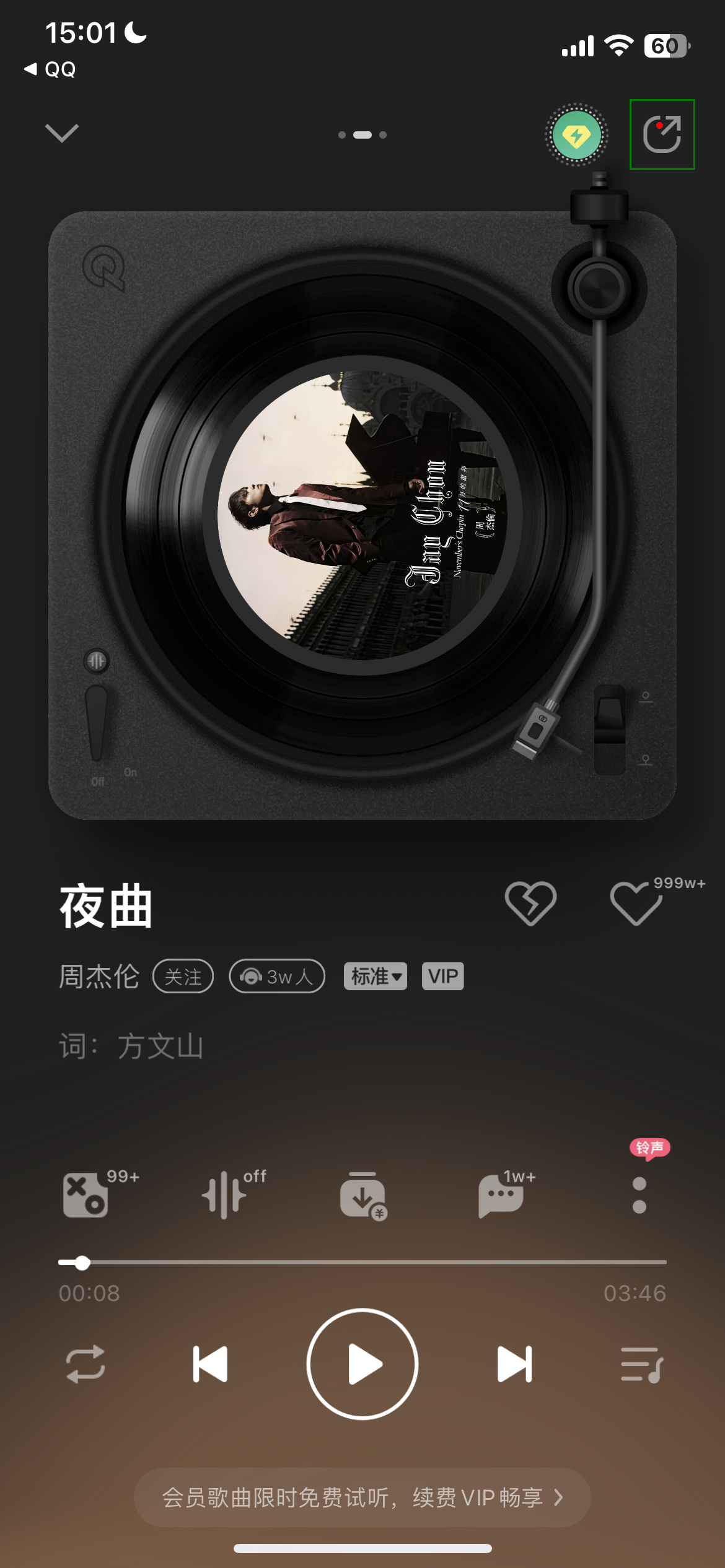}
        \caption{
        \textbf{\textcolor{darkgreen}{\cmark} Instruction: “Forwarding”.}
        \our~can identify what a forwarding button should look like.}
    \end{subfigure}
    \hfill
    \begin{subfigure}{0.24\textwidth}
        \centering
        \includegraphics[width=\textwidth]{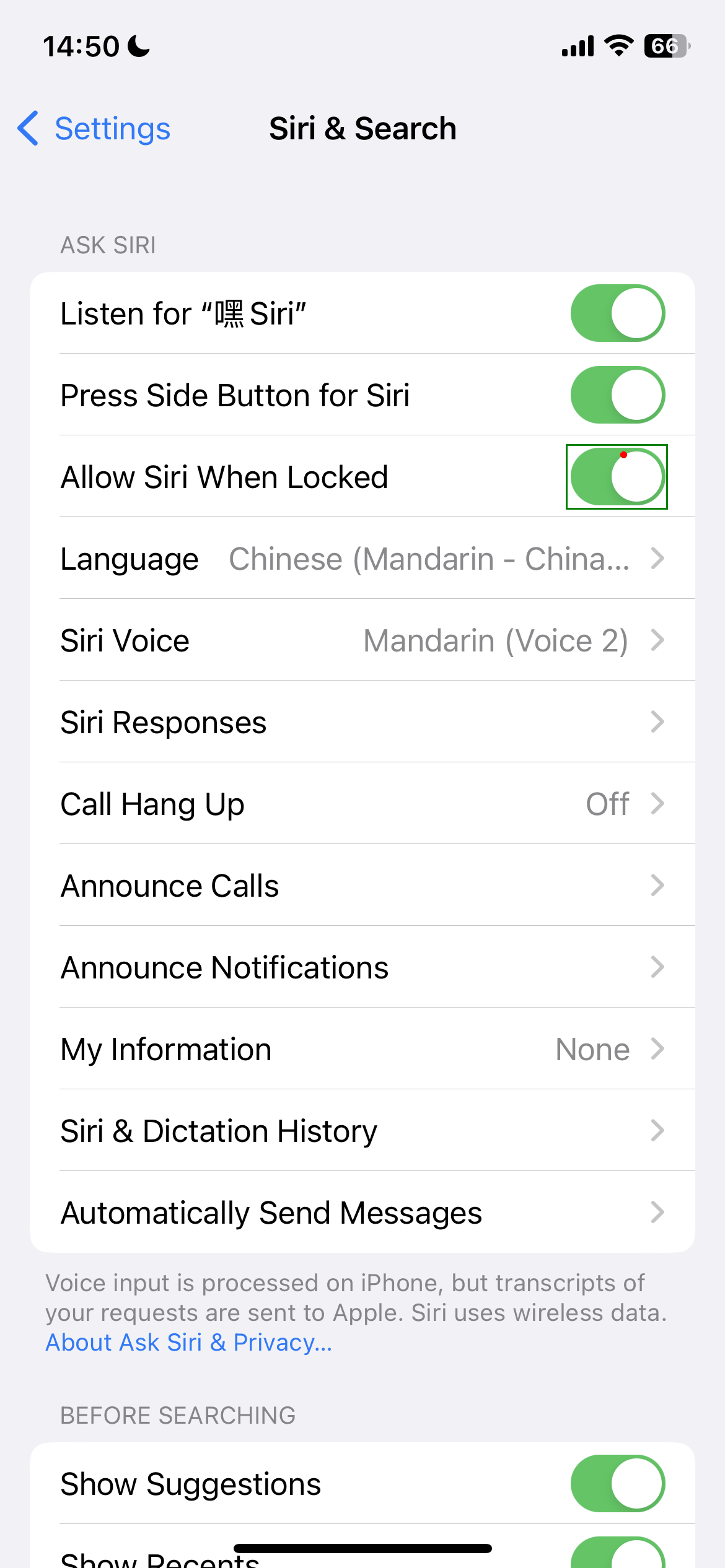}
        \caption{
        \textbf{\textcolor{darkgreen}{\cmark} Instruction: “Open allow siri when locked”.}
        \our~identifies the clickable element instead of the text itself.}
    \end{subfigure}
    \hfill
    \begin{subfigure}{0.24\textwidth}
        \centering
        \includegraphics[width=\textwidth]{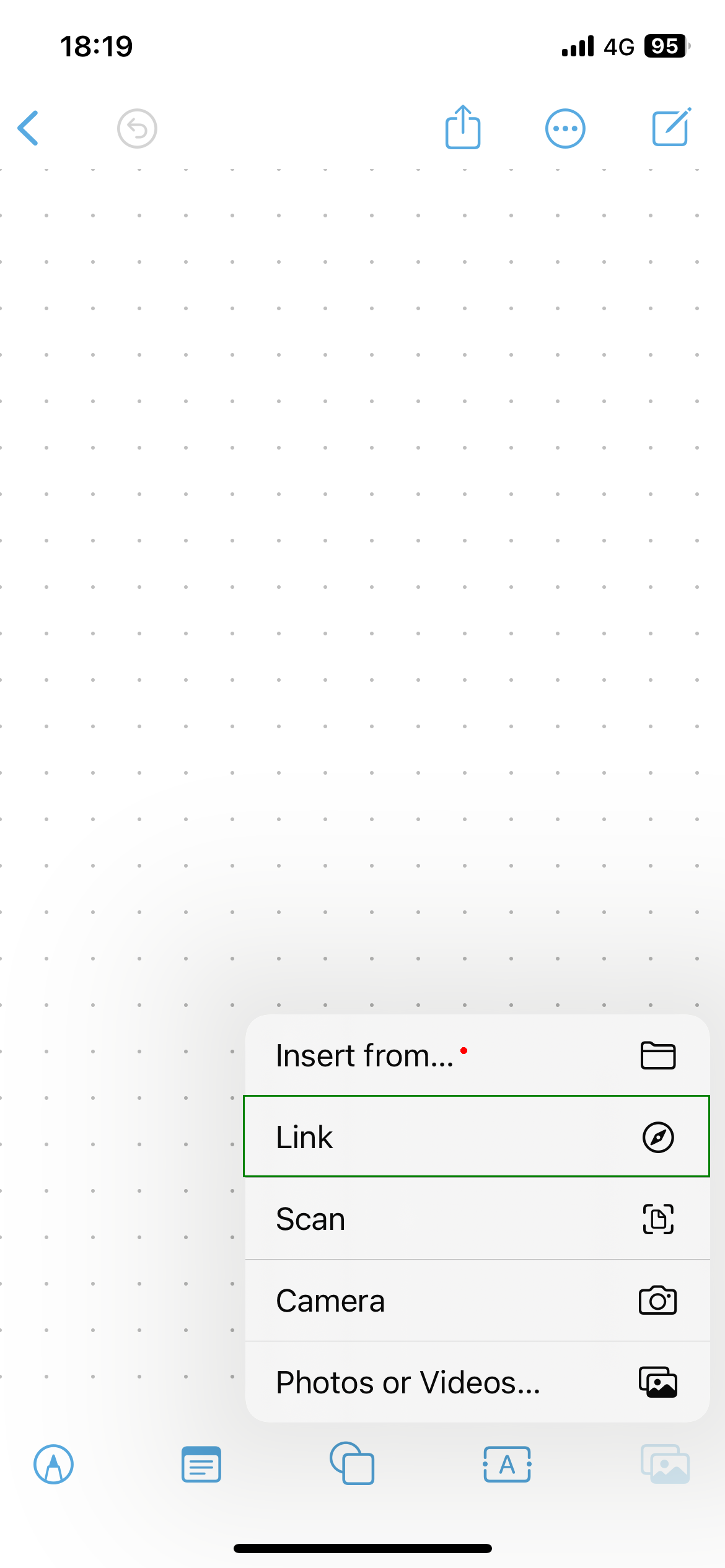}
        \caption{
        \textbf{\textcolor{darkred}{\xmark} Instruction: “Insert from link”.}
        The query being confusing as it contain both “Insert from” and “link”}
    \end{subfigure}
    \hfill
    \begin{subfigure}{0.24\textwidth}
        \centering
        \includegraphics[width=\textwidth]{figures/screenspot/False_mobile_text_screenspot_452_show_softwares.png}
        \caption{
        \textbf{\textcolor{darkred}{\xmark} Instruction: “Show softwares”.}
        The screenshot includes two software interfaces, causing confusion for the model.}
    \end{subfigure}
    \vspace{2em}
    \caption{Case studies on Screenspot's Mobile grounding.
    }
    \label{supp:fig:screenshot:mobile}
\end{figure*}

In Fig.\ref{supp:fig:screenshot:desktop} and \ref{supp:fig:screenshot:mobile}, we display several examples on Screenspot zero-shot grounding. We found that: 
with instruction tuning, \our~is able to perform some visual reasoning, such as it can distinguishes the correct operator among multiple abstract symbols or associate ‘view help’ with question mark, as shown in Fig.\ref{supp:fig:screenshot:desktop} (b,e).
Beside, we found in several failure cases, such as Fig.\ref{supp:fig:screenshot:desktop} (d,f), there might have multiple possible clickable elements, leading to model confusion.

\section{Related Works}
\noindent\textbf{GUI Agents.}
Recent studies reveal the potential of LLMs beyond language modeling, with advancements in~\cite{chain_of_thought, react, zeng2023socratic, yang2024gpt4tools, assistgpt} demonstrating their ability to autonomously complete complex tasks using tool integration. This has prompted GUI automation approaches such as:
{(i) Training-free systems}, which operate in separate stages: first, they gather UI information by converting the GUI to HTML representations, accessing the accessibility tree, or using visual models like OCR~\cite{pix2struct} and SoM~\cite{gpt4vsom, mmnav, zheng2024seeact}. Next, LLMs integrate this information to generate responses. This approach relies heavily on closed-source LLMs~\cite{gpt4}, resulting in high costs and limited applications, as real-world users generally only perceive screenshots rather than these oracle data sources.
To address these limitations, {(ii) Training-based models}~\cite{cogagent, seeclick, you2024ferret, li2024ferret} are proposed, which are pre-trained on large-scale UI vision-text corpora (e.g., screenshots), closing the visual perception gap to enable abilities like element grounding or navigation. 
Our work falls into the second category, focusing on key challenges in developing GUI visual agents, such as {efficiently modeling high-resolution visual screenshots} and {utilizing interleaved streaming for history management}.

\noindent\textbf{Vision-Language-Action Models.}
Vision-Language Models have recently made significant advancements, processing both visual and textual data, functioning effectively as flexible chatbots.
However, they still face limitations when require interaction with the real-world environment.
To address this, Vision-Language-Action (VLA) models have been developed to enhance physical or digital environmental perception and action generation.
Notable examples include RT-2-X~\cite{padalkar2023open} and OpenVLA~\cite{kim2024openvla}, which enable VLMs to perform actions in natural settings such as robotics~\cite{rt2,a3vlm,3dvla}, autonomous driving~\cite{covla}, and gaming~\cite{steve, omnijavis}.
While substantial progress has been made in  embodied contexts, the integration of VLA models into digital GUIs, remains less explored. 
In our work, we develop Vision-Language-Action (VLA) models for digital GUI environments to bridge this gap, focusing on the unique properties of GUIs and addressing how to model actions alongside other modalities within our VLA framework.


\paragraph{Efficient Visual Representations.}
The computational cost of Large Multimodal Models is a significant bottleneck, especially when scaling sequence lengths with high-resolution screenshots. 
The straightforward methods to reduce training costs, such as token pruning \cite{llava-prumerge
,fastv} and token merging \cite{chatunivi,llamavid}, are not suitable for GUI scenarios because GUIs require fine-grained location information rather than high-level semantics. These techniques discard essential spatial details, thereby impairing element grounding. Alternatively, Mixture-of-Depth (MoD)~\cite{mod,videollm-mod} offers a promising approach by allocating computation across model depths for language tokens, balancing performance with speed, and preserving the positional relationships of individual tokens.

In our work, we develop a UI-friendly solution for efficient visual token selection. 
High-resolution UI screenshots often contain redundant areas, like excessive whitespace, while maintaining a structured layout. To leverage this, we construct a UI connected graph in RGB space by treating each patch as a node and finding connected components to capture redundancy relationships. This graph guides self-attention block to skip part of redundant tokens, reducing computational costs without extra learnable parameters.
\section{Conclusions}

We introduced \our, a vision-language-action model for GUI visual agents that addresses key challenges in UI visual and action modeling, and instruction-tuning data curations. 
{From the model aspect}, our development of UI-Guided Visual Token Selection allows for efficient processing of high-resolution UI screenshots, significantly reducing computational costs. Our Interleaved Vision-Language-Action Streaming framework effectively manages complex interactions across modalities. 
{From the data aspect}, with a carefully curated, high-quality instruction-following dataset, \our achieves strong performance, with a lightweight model size.
These results demonstrate \our's potential to advance GUI visual agents towards more human-like interaction and perception.

\noindent\textbf{Limitation and Future work.}
Our model is primarily trained on offline data. A promising future direction is to enhance it in online environments through reinforcement learning, enabling deeper exploration of its limitations.
\appendix
\section{Datasets details}
\subsection{Instruction-Tuning data}
\noindent\textbf{Website.}
We develop a parser using PyAutoGUI~\cite{pyautogui} and source websites from 22 representative scenarios such as Airbnb, Booking, AMD, and Apple, which covering shopping, technology, etc. 
For each scenario, we collect multiple screenshots to maximize annotation coverage.
This process yields 22K screenshots with a total of 926K element annotations. After filtering out elements classified as \texttt{static text}, we retain 576K elements, averaging 26 elements per screenshot.

\noindent\textbf{Mobile.}
We source mobile data from AMEX~\cite{amex} annotations: (i) Element grounding and (ii) Functionalities.
which covers 97K screenshots, with 885K elements and 178K functionalities.

\noindent\textbf{Desktop.}
We collect 100 screenshots and 2,000 raw annotations from the OmniAct~\cite{omniact} Desktop training split, encompassing 15 software applications across iOS, Windows, and Linux desktops. Additionally, we augment these annotations using GPT-4o-assisted prompting, as detailed in the following section.

\subsection{Downstream tasks}
\noindent\textbf{Mind2Web}~\cite{mind2web} supports the development of generalist web agents capable of completing complex tasks on any website by following language instructions. The dataset aligns each HTML document with its corresponding webpage screenshot, featuring a training set of 7,775 actions and three test splits—test-task, test-website, and test-domain—verified for correct rendering and element visibility to ensure reliable evaluation across tasks, websites, and domains.
It action space includes three actions: \texttt{CLICK}, \texttt{TYPE} and \texttt{SELECT}.

\noindent\textbf{AITW}~\cite{aitw} is an Android smartphone environment, which contains 30k instructions and 715K trajectories.
We follow the setting by SeeClick~\cite{seeclick}, which divide the data by domains: General, Install, Google Apps, Single, Web Shopping.
It action space includes 12 actions: \texttt{CLICK}, \texttt{TYPE}, \texttt{SELECT}, \texttt{SCROLL UP}, \texttt{SCROLL DOWN}, \texttt{SCROLL LEFT}, \texttt{SCROLL RIGHT}, \texttt{PRESS BACK}, \texttt{PRESS HOME}, \texttt{PRESS ENTER}, \texttt{STATUS TASK COMPLETE}, \texttt{STATUS TASK IMPOSSIBLE}.

\noindent\textbf{MiniWob}~\cite{miniwob++} comprises 2000 open-ended tasks from 137 real web environments, each with high-level instruction and action trajectory. It action space includes 2 actions:  \texttt{CLICK}, \texttt{TYPE}.

\section{Settings}

\subsection{Training details}
We utilize 32 V100 GPUs for instruction-tuning, while downstream adaptation is conducted on 8 V100 GPUs. 
The batch size per GPU is set to 1, with gradient accumulation steps of 2. We use float16 precision for training. To enhance efficiency, we apply LoRA tuning with a rank of 64 and an alpha value of 128 for both the language model and visual encoder, resulting in 4\% of the total learnable parameters. 
We leverage DeepSpeed Zero-2 and use the SDPA attention mechanism.
The learning rate is configured  to 1e-4.
The maximum visual patch number is $1280$. 
The instruction-tuning training duration is approximately two days.

\noindent\textbf{UI Connected Graph:}
We apply the UI-Graph to both the visual encoder and language model with a masking ratio of 0.75, using cross-layer insertion at layer 14. During each iteration, a random ratio of visual tokens is masked. For inference usage, we uniformly sample tokens across each component, ensuring visibility of all components to the model.

\noindent\textbf{Interleaved Streaming:}
In our streaming setting, we set up the history number as 2.

\noindent\textbf{Data Resampling:} To achieve data balance among existing datasets, probabilities are assigned to each dataset, with weights set as (Web:Mobile:Desktop: GUIAct-Web:GUIAct-Mobile) = (1:1:1: 1:1).

\subsection{Prompt templates}
\noindent\textbf{GPT-4o Assisted Prompts.}
Below, we display the prompt used for GPT-4o to augment the OmniAct original annotations, which mainly includes three types: 
(i) Appearance; (ii) Spatial-relationship; (iii) Situational.
\begin{tcolorbox}[boxrule=0pt, colframe=white, sharp corners, left=1mm, right=1mm, top=0.2mm, bottom=0.2mm]
\footnotesize
{
You will receive a screenshot with a red bounding box surrounding the target element, along with the target element’s name. 
Analyze the screenshot and provide concise descriptive responses for each of the following dimensions.
\\\\
\textbf{1.Appearance:} Describe the target element’s color, shape, ocr and other visual characteristics.\\
   - Example: “A rectangular chat card with a blue background to ‘Ash’.”

\textbf{2.Spatial:} Describe the target element’s position based on the contextual spatial relationship.\\
   - Example: “The element that positioned above the Clara.”

\textbf{3.Situational:} Create an intent-oriented query related to the target element, considering how a user might interact with it.\\
   - Example: “Send a message to Ash.”
\\\\
Please follow these guidelines:\\
- Do not confuse the red bounding box with the element itself.\\
- Provide responses as concise sentences (15 words or fewer).\\
- For each types, make the description specific enough to distinguish it from other elements.\\
- If a dimension does not apply, respond with "None."\\
- Structure your response in JSON format as shown below:
\\\\
\texttt{
\{\\
"appearance": "A rectangular chat card with a blue background with letter 'A'",\\
"spatial": "The element that positioned above the Clara.",\\
"situational": "Send a message to Ash."\\
\}
}
}
\end{tcolorbox}

\noindent\textbf{Action README Template.}
Below, we present the template for action navigation. Variables, marked in \textcolor{NiceBlue}{\texttt{Blue}}, depend on specific scenarios, while actions used for loss calculation are highlighted in \textcolor{red}{\texttt{Red}}.

\begin{tcolorbox}[boxrule=0pt, colframe=white, sharp corners, left=1mm, right=1mm, top=0.2mm, bottom=0.2mm]
\footnotesize
{
\textcolor{black}{You are an assistant trained to navigate the }
\textcolor{NiceBlue}{\texttt{\{device\}}} 
\textcolor{black}{
. Given a task instruction, a screen observation, and an action history sequence, 
output the next action and wait for the next observation.\\}

\textcolor{black}{Here is the action space:
}

\textcolor{NiceBlue}{
\textit{\# templated by action\_type with action description.}\\
1. ‘CLICK’: Click on an element, value is the element to click and the position [x,y] is required.\\
2. ‘TYPE’: Type a string into an element, value is the string to type and the position [x,y] is not applicable.\\
...
\\
}

\textcolor{black}{Format the action as a dictionary with the following keys:\\
{\{‘action’:‘action\_type’, ‘value’:‘element’, ‘position’:[x,y]}\}\\
Position represents the relative coordinates on the screenshot and should be scaled to a range of 0-1.\\}

\textcolor{black}{Task: }\textcolor{NiceBlue}{\texttt{\{task\}}}


\textcolor{NiceBlue}{\texttt{<past\_image\_1>}\textcolor{red}{\texttt{\{past\_action\_1\}}}
\\
...\\
\texttt{<past\_image\_n>}\textcolor{red}{\texttt{\{past\_action\_n\}}}
\textcolor{NiceBlue}{\texttt{<image\_{n+1}>}}\textcolor{red}{\texttt{\{action\_n+1\}}}
}
}
\end{tcolorbox}

{\small
\bibliographystyle{ieee_fullname}
\bibliography{main}
}

\end{document}